\documentclass[lettersize,journal]{IEEEtran}

\usepackage{amsmath,amsfonts}
\usepackage{array}
\usepackage{textcomp}
\usepackage{stfloats}
\usepackage{url}
\usepackage{verbatim}
\usepackage{graphicx}
\usepackage{wrapfig}
\usepackage{booktabs}
\usepackage{bbding}
\usepackage{xcolor}
\usepackage{multirow}
\usepackage{enumitem}
\usepackage[utf8]{inputenc}
\usepackage[T1]{fontenc}
\usepackage{nicefrac}
\usepackage{microtype}
\usepackage{varwidth}
\usepackage{makecell}
\usepackage{colortbl}
\usepackage[most]{tcolorbox}
\usepackage{soul}
\usepackage{adjustbox}
\usepackage{orcidlink}
\usepackage{amssymb}

\usepackage{cite}

\usepackage{subcaption}  

\usepackage{caption}
\usepackage{capt-of}

\usepackage[ruled,vlined]{algorithm2e}

\usepackage{hyperref}

\makeatletter
\@ifundefined{c@lofdepth}{\let\c@lofdepth\relax}{}
\@ifundefined{c@lotdepth}{\let\c@lotdepth\relax}{}
\makeatother

\usepackage{tocloft} 

\definecolor{ForestGreen}{RGB}{34,139,34}

\begin{document}

\title{Don’t Pause: Streaming Video-Language Synchrony for Online Video Understanding} 

\author{Zhenyu Yang\,\orcidlink{0009-0005-5298-0543}, Kairui Zhang\,\orcidlink{0009-0000-9155-0988}, Shengsheng Qian\,\orcidlink{0000-0001-9488-2208},~\IEEEmembership{Member,~IEEE,} Weiming Dong\,\orcidlink{0000-0001-6502-145X},~\IEEEmembership{Member,~IEEE,} and Changsheng Xu\,\orcidlink{0000-0001-8343-9665},~\IEEEmembership{Fellow,~IEEE}


}

\markboth{Preprint, 2026}
{Yang \MakeLowercase{\textit{et al.}}: Don't Pause: Streaming Video-Language Synchrony for Online Video Understanding}

\IEEEpubid{}

\maketitle

\begin{abstract}
Online Video Large Language Models (Video-LLMs) have advanced toward seamless human-AI interaction through frame-by-frame processing and proactive responding. However, a critical challenge remains in streaming scenarios: existing models typically pause video perception while generating responses, breaking real-time video-language synchrony and causing stutters. To address this, we introduce a novel paradigm for online video understanding: Streaming Video-Language Synchrony (SVLS), and present LyraV, a live streaming assistant built upon a hierarchical control framework with two core innovations. First, the Frame-Driven Transition Controller (FDTC), a training-free verification-based finite-state machine, makes high-level semantic decisions on when to continue speaking, start a new response, or stay silent. Second, the Streaming Token Pacer (SToP), a plug-and-play lightweight predictive module, dynamically adapts the language generation rate to match the pace of the visual content. Concretely, LyraV performs \emph{per-frame incremental, sub-budget decoding}: within each frame interval it emits only a small chunk of tokens that fits the real-time budget, so perception is never blocked for a full sentence. Together, these components enable LyraV to seamlessly interleave incoming video frames with generated word tokens, achieving a fine-grained synchrony. Extensive experiments conducted on five online and three offline benchmarks demonstrate that LyraV preserves the backbone's general understanding ability while substantially improving streaming synchrony and narrative fluency, delivering a 98.29\% synchrony with video playback and a real-time processing speed of 3.89 FPS. Interestingly, we observe an empirical capability in LyraV: dynamic reasoning over streaming tokens, enabling continuous interpretation and "thinking" alongside visual input. 
\end{abstract}

\begin{IEEEkeywords}
Online video understanding, streaming video models, proactive response.
\end{IEEEkeywords}

\section{Introduction}
\label{sec:intro}

\IEEEPARstart{A}{chieving} human-like response capabilities for online video assistants represents a critical frontier in AI research. Recent advances in Video Large Language Models (Video-LLMs)~\cite{ataallah2024minigpt4, maaz2023video,li2023videochat,yang2023vid2seq,wang2022internvideo,11050020,11223149} have significantly advanced real-world human-AI interaction through innovations in spatiotemporal perception~\cite{cheng2024videollama, liu2024oryx, ren2024timechat,11202655}, memory-augmented compression~\cite{zhang2024flash, song2024moviechat+, he2024ma}, and long-term video understanding~\cite{wang2024longllavascalingmultimodalllms, longvila, zhang2024long, 11359544,11430664,11535730}. Meanwhile, emerging platforms (including smart glasses~\cite{lee2018interaction}, head-mounted devices~\cite{sutherland1968head}, and robotic assistants~\cite{horn1986robot}) urgently require J.A.R.V.I.S.-like AI systems capable of continuous multimodal perception and interaction. Such assistants must process perpetual sensory input while remaining always-on to facilitate seamless user engagement in a streaming paradigm.

\begin{figure*}[t]
\centering
\includegraphics[width=1.0\textwidth]{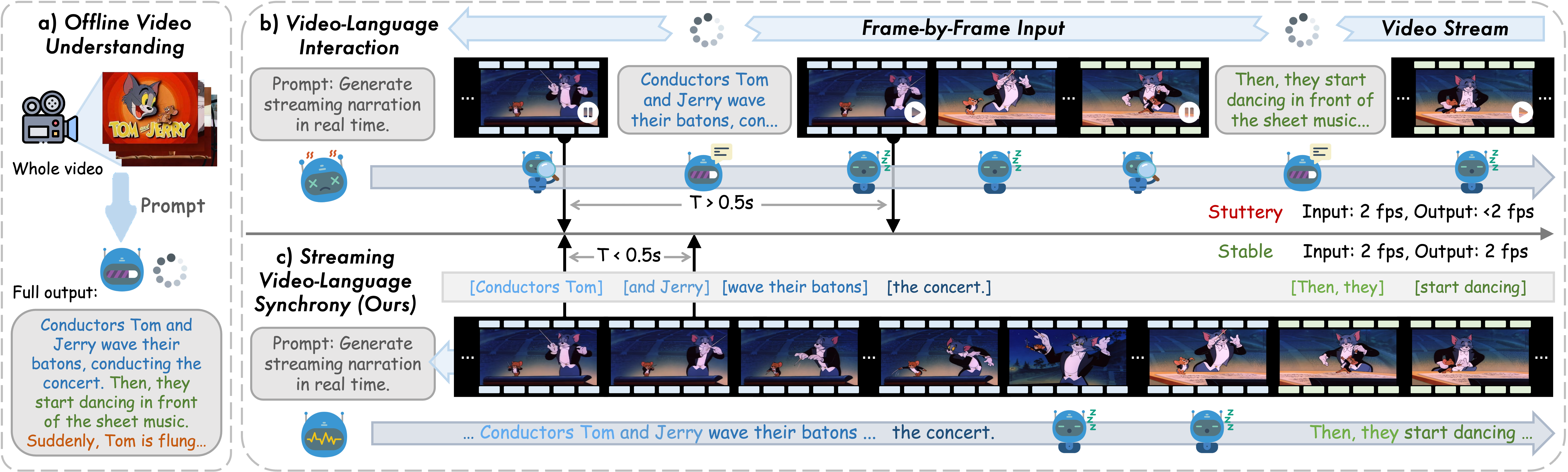} 
\caption{\small \textbf{Illustration of Streaming Video-Language Synchrony.} (a) Traditional offline methods generate responses only after processing the entire video. 
(b) Existing online methods often pause video perception to decode full sentences, breaking real-time synchrony and causing stutters. (c) Our proposed SVLS paradigm, enabled by LyraV, achieves concurrent perception and generation. It seamlessly interleaves incoming video frames with generated word tokens, maintaining continuous alignment.}
\label{fig:overview}
\end{figure*}

To address this need, researchers are pushing to adapt offline Video-LLMs into online video assistants~\cite{chen2024videollm, wu2024videollm, li2025lion, ding2025streammind,wang2024videollm,wang2025streambridge,qian2024streaming} (Fig. \ref{fig:overview}(a)→(b)), which process frame-wise inputs, reason through time, and deliver proactive responses.
Recently, three primary approaches have emerged: (1) \textit{response gating}, (2) \textit{end-of-sequence (EOS) token prediction}, and (3) \textit{verification-based} methods. Response gating methods~\cite{ding2025streammind,wang2024videollm,wang2025streambridge}, exemplified by MMDuet~\cite{wang2024videollm}, train a binary classifier to decide between generating a response or remaining silent at each step. The token-based methods~\cite{chen2024videollm,wu2024videollm,li2025lion}, proposed in VideoLLM-online~\cite{chen2024videollm}, introduce an EOS prediction mechanism to dynamically determine when to keep silent by generating only an [EOS]. Finally, verification-based methods (e.g., LiveStar~\cite{yang2025livestar}) determine the optimal timing for proactive responses by using previously decoded outputs within a single forward pass, effectively mitigating repetitive or uninformative outputs.

However, current online Video-LLMs essentially \textbf{pause} the video during playback to decode responses and resume only after generation completes, failing to achieve human-level video-language synchrony (i.e., perceiving while speaking). Existing models merely interleave full-sentence language outputs within visual streams. In contrast, humans seamlessly integrate and interleave perception with speech without interrupting sensory input (i.e., continuous video streams). As illustrated in Fig. \ref{fig:overview}(c), models should process incoming frames \textit{while} emitting text tokens at frame-level granularity, forming a unified stream of interleaved video frames and words/chunks to achieve streaming video-language synchrony. We use ``concurrent'' throughout in this operational sense---perception is never paused for an entire sentence, and at most a small per-frame token chunk is decoded within each frame budget---rather than implying literally simultaneous perception and generation within a single instant. 
Detailed cases are shown in Fig.~\ref{fig:case_study_lyrav_1}.
While systems like SpeakStream~\cite{bai2025speakstream} incrementally generate audio from streaming text using decoder-only architectures, video-language frameworks lack exploration of streaming frame-word interleaving. Motivated by this critical gap, we address \textbf{Challenge 1:} \textit{How to achieve concurrent streaming perception and generation for seamless video-language synchrony?}

Moreover, beyond fine-grained synchrony, determining optimal response timing remains crucial for online video assistants. While existing response-silence strategies address proactive output in traditional online tasks, streaming video-language synchrony introduces a new hurdle: at each new frame, the model must decide whether to \textit{continue} incomplete utterances or \textit{trigger} new responses. For example, in live sports commentary~\cite{giancola2018soccernet,taniguchi2019generating} ("He shoots... it's heading towards... it's in! GOAL!... Wait—no! The keeper saves it!"), rapidly evolving events trigger responses that interrupt prior narration. Conversely, in sparse-event scenarios (e.g., descriptive captioning~\cite{chen2024sharegpt4video,xu2024pllava,islam2024video}), models must sustain narration within the same event clip until completion, remaining silent during subsequent frames of that event.  Crucially, since existing online Video-LLMs pause visual processing to decode responses, generating outputs only at discrete points, they inherently cannot resolve this dynamic arbitration problem. This limitation motivates \textbf{Challenge 2:} \textit{How to dynamically arbitrate between utterance continuation and new response triggering?}

To address these challenges, we introduce LyraV, an innovative live streaming assistant that establishes a new paradigm for online video understanding: Streaming Video-Language Synchrony (SVLS), as illustrated in Fig.~\ref{fig:overview}(c). Rather than a new end-to-end model, LyraV is realized as a lightweight, plug-and-play synchrony control layer wrapped around a frozen online Video-LLM backbone, through two components that work in concert.
To solve \textbf{Challenge 1}, the fine-grained task of concurrent perception and generation, we propose the Streaming Token Pacer (SToP), the only trainable module in our framework. This decoding framework adaptively optimizes the number of tokens emitted per frame under real-time latency constraints, ensuring language generation is seamlessly interleaved with the video stream without buffering pauses.
To address \textbf{Challenge 2}, the high-level arbitration of narrative timing, we present the Frame-Driven Transition Controller (FDTC), a training-free control mechanism. It leverages a verification-based finite state machine to dynamically decide whether to continue an ongoing utterance, trigger a new response, or remain silent, based on the evolving visual context. Extensive experiments on five online and three offline benchmarks show that LyraV preserves the backbone's general video understanding while delivering state-of-the-art streaming synchrony and narrative fluency.

Our contributions can be summarized as follows:
\begin{itemize} 
\item 
We fundamentally rethink video-language interaction by introducing the novel paradigm of Streaming Video-Language Synchrony (\textbf{SVLS}), enabling Video-LLMs to achieve concurrent streaming perception and generation, seamlessly interleaving video frames and word tokens without requiring long pauses for decoding.

\item 
We present \textbf{LyraV}, an online video understanding assistant. Its core innovation is the Frame-Driven Transition Controller (\textbf{FDTC}), which performs state transitions and determines at every video frame whether to continue the current utterance, trigger a new response, or remain silent, based on evolving visual context.

\item 

We propose Streaming Token Pacer (\textbf{SToP}), a decoding framework that adaptively optimizes the token number emitted per frame under real-time latency constraints. By adjusting the speaking rate to match the pacing of visual events, SToP enables stutter-free, human-like video-language streaming without buffering pauses.

\item
Extensive experiments conducted on 8 benchmarks show that LyraV attains state-of-the-art streaming synchrony and narrative fluency under real-time constraints, while preserving the general ability. We further qualitatively observe a dynamic reasoning behavior, where LyraV incrementally refines its narration as the video unfolds.

\end{itemize}

\section{Related Work}
\label{sec:related_work}

\subsection{Offline Video Large Language Models.}
Advances in Large Language Models (LLMs)~\cite{touvron2023llama,team2023gemini,achiam2023gpt,ouyang2022training,radford2018improving, yang2024ldre} has catalyzed a surge of research interest in Video Large Language Models (Video-LLMs)~\cite{ataallah2024minigpt4, maaz2023video,li2023videochat,yang2023vid2seq,wang2022internvideo, lin2024vila}, especially for improving video understanding tasks (such as video captioning~\cite{chen2024sharegpt4video,xu2024pllava,islam2024video, yang2024semantic}, video question answering~\cite{ko2023large,li2024mvbench,maaz2024videogpt+, zhang2024llavanext-video,9770842,10214041,11506215,11329152,11219357,11391656,11146594}, and temporal grounding~\cite{guo2025vtg,xu2024vtg,wang2024hawkeye,11184436}) in offline scenarios.
Alongside widely-used proprietary models like GPT-4V~\cite{yang2023dawn}, GPT-4o~\cite{achiam2023gpt}, and Gemini 2.5 Pro~\cite{comanici2025gemini, reid2024gemini}, a growing number of open-source Video-LLMs~\cite{10721284,10670217,10815073,10839067}, including Videollama 2~\cite{cheng2024videollama}, Video-LLaVA~\cite{lin2023video}, and VideoChat2~\cite{li2024mvbench}, have also shown remarkable performance in these areas.
Effective operation in streaming settings~\cite{chen2024videollm,qian2024streaming} must address two core challenges: first, the limited context window of models like GPT-4V~\cite{yang2023dawn} and GPT-4o~\cite{achiam2023gpt}, which restricts input to a small number of frames; and second, the inability to process complete videos, requiring instead the real-time, frame-by-frame processing of online streams with temporal awareness to respond at appropriate moments. To this end, we propose LyraV, which processes perpetual sensory input while remaining always-on, thereby enabling seamless user engagement within a streaming paradigm.

\subsection{Online Video Large Language Models.}
Recent years have witnessed increasing attention to online video understanding~\cite{qian2024streaming, chen2024videollm, zhou2024streaming, xiong2025streaming}, motivated by the proliferation of diverse streaming video sources such as online platforms~\cite{zellers2022merlot}, live streaming services~\cite{gao2023livechat}, and wearable cameras~\cite{grauman2022ego4d}.  
In contrast to passively responsive models~\cite{zhou2024streaming, xiong2025streaming}, three main strategies have arisen for proactive response generation: (1) response gating~\cite{fu2025vispeak, kang2025open, yang2025streamagent, qianlearning, kim2025egospeak, mun2019streamlined, yan2026proact, azad2026streamready, tian2026roma, kim2026stride, zheng2026garde, yao2025timechat}, which trains a binary classifier to decide between responding or staying silent at each step, as in StreamMind~\cite{ding2025streammind}; (2) EOS token prediction~\cite{chen2025livecc,zhang2025eyes, zhang2025proactive, yang2025assistpda, xia2025streaming, wang2025mmduet2, liu2026thinking, chen2026streamingclaw}, introduced by VideoLLM-online~\cite{chen2024videollm}, where an EOS token is generated to indicate silence dynamically; and (3) verification-based techniques~\cite{zhang2026querystream,yang2025livestar}, such as LiveStar~\cite{yang2025livestar}, which leverage previously decoded outputs in a single forward pass to identify the best response timing and reduce redundant outputs.  
However, they typically halt video streams during text generation, disrupting the synchrony between visual perception and language output—unlike humans, who fluidly interleave seeing and speaking. To overcome it, we present LyraV, which incorporates a streaming token pacer to dynamically optimize token emission per frame under real-time constraints, enabling simultaneous video processing and text generation without buffering delays.

\section{Paradigm: Streaming Video-Language Synchrony} %

\textbf{Offline Video Understanding.}
Consider a video $V$ comprising $K$ frames, denoted as $\{[Frm^{t_i}]\}_{i=0}^{K-1}$. Taking the dense video captioning task as an example, we define a VideoLLM $\psi(\cdot)$. Given the entire video $V$ as input, $\psi$ generates a caption at each designated decoding point within a predefined set $P \subseteq \{t_i\}_{i=0}^{K-1}$. This model operates end-to-end: it processes the complete video input and outputs the dense captions.
The process is formally expressed as:
\begin{equation}
\psi(V=\{[Frm^{t_i}]\}_{i=0}^{K-1}) = \big\{ \langle t, [R s p]^{t} \rangle \mid t \in P \big\},
\end{equation}
where $[Rsp]^{t}$ represents the generated caption at time $t$.

\noindent \textbf{Traditional Video-Language Interaction for Online Video Understanding.}
Online video understanding requires frame-by-frame processing and proactive response generation, necessitating finer-grained video-language interaction where sentences and video clips are interleaved. For comparative formulation simplicity, we consider a video segment with a single response time $t_p$.
The model $\psi$ processes frames sequentially and generates output as:
\begin{align}
\psi([Frm^{t_i}]) = 
\begin{cases} 
\{[Rsp]_j\}_{j=0}^{N-1}, &\text{if } i = p \\ %
\ \ \ \ \ \ \ \ \ \varnothing, &\text{otherwise}
\end{cases}
\end{align}
where $\varnothing$ denotes silent processing (no output), while $\{[Rsp]_j\}_{j=0}^{N-1}$ represents the complete decoded sentence of $N$ tokens at response time $t_p$.

\noindent \textbf{Streaming Video-Language Synchrony (SVLS).}
To achieve concurrent streaming perception and generation, seamlessly interweaving video frames and word tokens, SVLS outputs $m$ word tokens per frame after the triggering time $t_p$, rather than generating the full sentence at once. 
Note that $m$ can be dynamically adjusted \textit{per frame} based on instantaneous computational load and latency requirements.  
Formally, $\psi$ processes frames sequentially and generates:
\begin{align}
\label{eq:svls}
\psi([Frm^{t_i}]) = 
\begin{cases} 
\left\{ [Rsp]_j \right\}_{j = j_{\text{start}}}^{j_{\text{end}}}, & \text{if } p \leq i < i_{\max} \\
\ \ \ \ \ \ \ \ \ \ \varnothing, & \text{otherwise}
\end{cases}
\end{align}
where:
\begin{align*}
j_{\text{start}} &= (i - p) \cdot m, \\
j_{\text{end}} &= \min\left\{ (i - p + 1) \cdot m - 1,\; N-1 \right\}, \\
i_{\max} &= \min\left\{ \left\lceil N/m \right\rceil + p,\; K \right\}.
\end{align*}
Here $i_{\max}$ is the cutoff frame index for response generation.
For notational simplicity, Eq.~\ref{eq:svls} writes a constant per-frame budget $m$; in practice $m_i$ is predicted per frame and further capped by a hard latency cutoff (Sec.~\ref{sec:SVLS_experiments}), and each incoming frame first triggers a brief verification step before decoding. SVLS therefore does not claim instantaneous co-execution of perception and generation; rather, it guarantees that perception is never blocked for the duration of a full sentence, decoding at most a sub-budget token chunk per frame.
Thus, SVLS establishes a new paradigm for online understanding by synchronizing visual perception and language generation at the granularity of individual frames and tokens.

\begin{figure*}[t]
\centering
\includegraphics[width=0.95\textwidth]{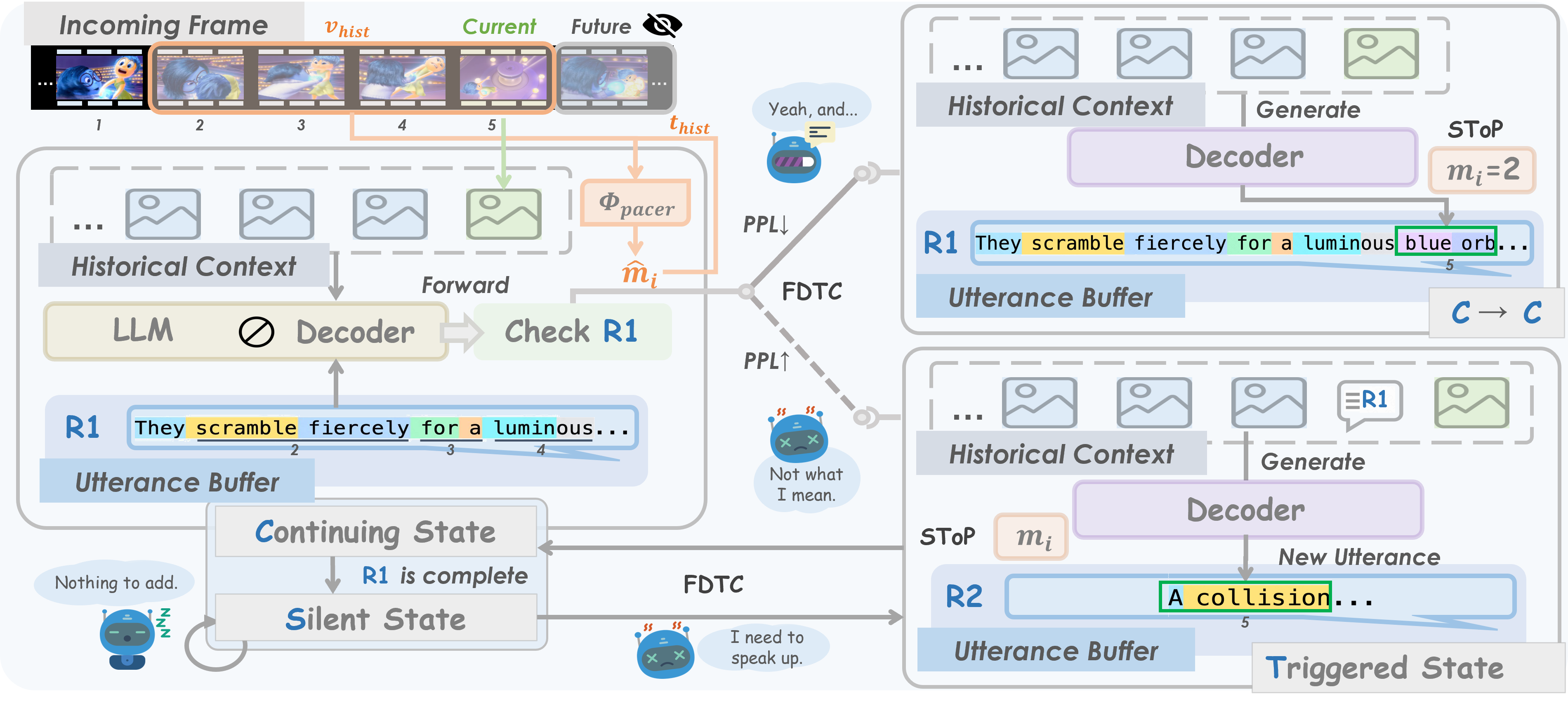} 
\caption{\small \textbf{Architecture of the proposed FDTC framework.} The Frame-Driven Transition Controller performs frame-by-frame analysis to regulate state transitions between Silent, Continuing, and Triggered modes, determining whether to continue the current utterance, trigger a new response, or remain silent, based on evolving visual context in real-time. For a clear state transition diagram, refer to Fig. \ref{fig:FDTC_SToP}(a).}
\label{fig:model_architecture}
\end{figure*}

\section{Method: LyraV}
\subsection{Framework Overview and Inference Pipeline}
LyraV achieves fine-grained SVLS through a hierarchical control framework that decouples the intricate decision-making process underlying streaming video narration.
It separates high-level semantic decisions ("\textit{when to initiate or conclude an utterance}") from low-level real-time controls governing the pacing of articulation ("\textit{how fast to articulate it}"). It enables LyraV to generate fluent and adaptive narrations that closely mirror the natural rhythm of human speech, all while operating seamlessly in a streaming video context.

\noindent \textbf{Model Architecture.} 
As illustrated in Fig.~\ref{fig:model_architecture}, LyraV implements a hierarchical control framework built upon three core components. The foundation is a frozen online Video-LLM backbone~\cite{yang2025livestar}, providing capabilities for online video understanding and language generation.
Built upon this backbone is a dual-tier control system that governs real-time streaming communicative behavior. The Frame-Driven Transition Controller (FDTC) acts as the high-level regulator of \textit{when to speak}. It manages a finite-state machine with three modes (Continuing, Triggered, and Silent), transitioning between them based on a verification mechanism that evaluates utterance perplexity (PPL) against incoming visual cues.
Complementing it, the Streaming Token Pacer (SToP) serves as a fine-grained controller that regulates \textit{how fast to speak}. It employs a lightweight Transformer encoder to predict a content-aware token count $\hat{m}_i$, followed by a latency cut-off mechanism that constrains the final token output $m_i$ within the real-time frame budget $\Delta t$.
Together, FDTC and SToP enable LyraV to achieve semantically coherent and temporally synchronized video-language streaming.

\noindent \textbf{Verification Mechanism and Streaming Context.}
A verification-based decoding mechanism is employed in LyraV for decision-making. Building on the general principle of verification-based decoding~\cite{yang2025livestar}, we extend it from a single-pass, full-caption check into an \emph{incremental, stateful} verification that dynamically evaluates semantic alignment between an \emph{ongoing} response prefix and the evolving video stream, without explicit silence tokens.
For streaming integration, we build a scene-local context where frames from the current scene are temporally grouped and concatenated with tokens from the ongoing response, forming $\mathcal{C}_t = [Frm_1, \dots, Frm_t; Rsp_{1:\ell}]$, where $Frm_1$ is the scene’s first frame and $Rsp_{1:\ell}$ are tokens generated so far. This allows the model in Continuing State to naturally extend the prior utterance using the accumulated visual prefix and preceding tokens.
At each new frame, LyraV computes PPL of $Rsp_{1:\ell}$ given the full visual sequence. A stable or decreasing PPL maintains the Continuing state, while a sharp increase indicates semantic drift, triggering FDTC to terminate the utterance and begin a new one, enabling adaptive, interpretable state control in streaming scenarios.

\noindent \textbf{Inference Pipeline.}
The inference process in LyraV follows a streamlined, frame-by-frame pipeline that seamlessly unifies its control modules. For each incoming frame $\text{Frm}_t$ ingested, it first updates the multimodal streaming context $\mathcal{C}_t$. FDTC then evaluates $\mathcal{C}_t$, using a verification mechanism to compute utterance PPL and determine the next model state: Continuing, Triggered, or Silent. If the state corresponds to text generation (i.e., Continuing or Triggered), SToP is activated to estimate the ideal token count $\hat{m}_t$ for the current frame. Finally, the foundational Video-LLM performs auto-regressive decoding, where the actual number of tokens $m_t$ is regulated by both $\hat{m}_t$ and the real-time latency budget $\Delta t$. 
This cycle repeats per frame, enabling LyraV to produce a fluid and temporally aligned narration stream. See Algorithm~\ref{alg:LyraStar} for the algorithm flowchart.

\subsection{Frame-Driven Transition Controller}
\label{sec:FDTC}

\begin{figure}[t]
    \centering
    \includegraphics[width=\linewidth]{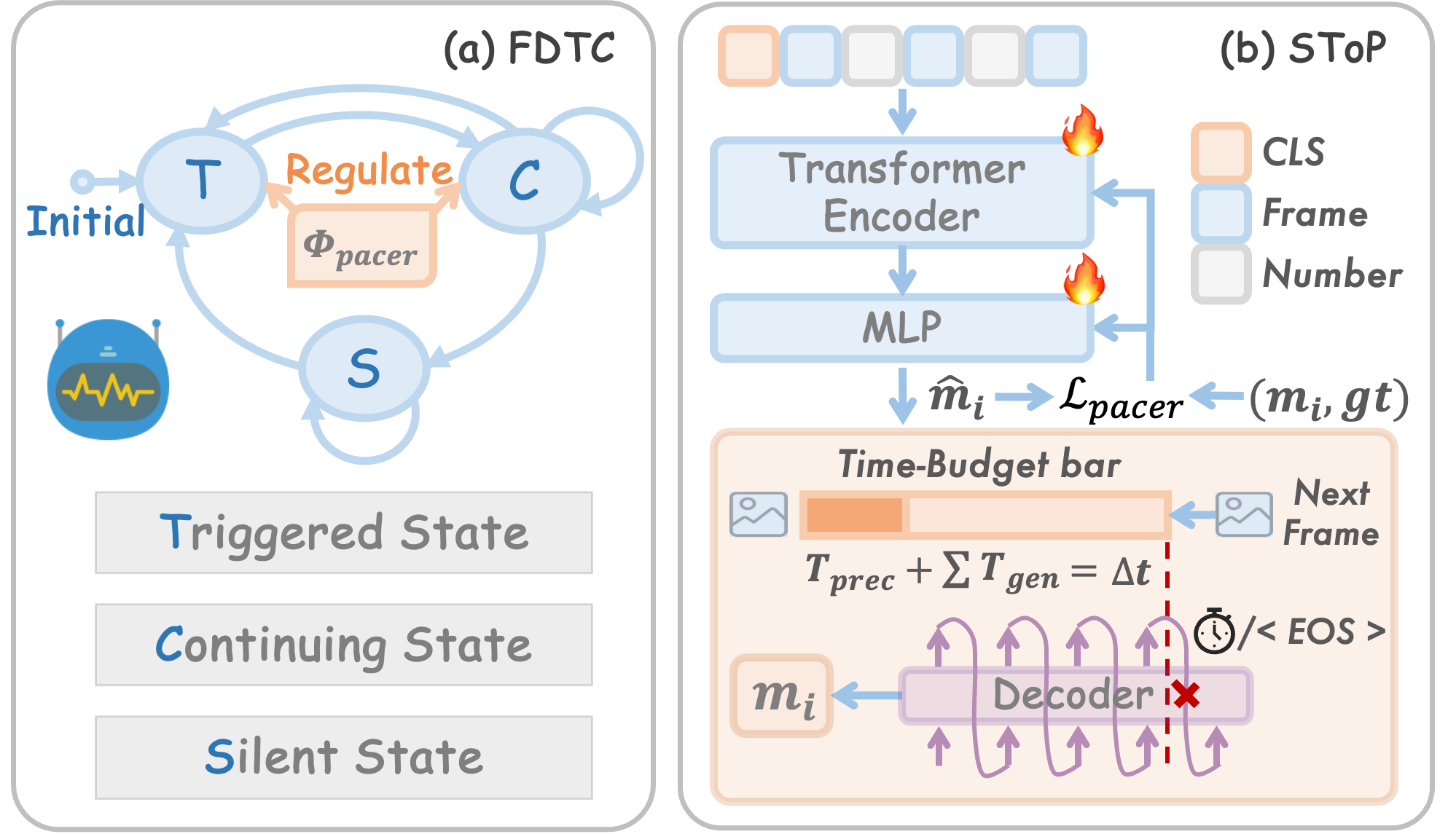}
    \caption{(a) \textbf{Overview of FDTC:} A frame-driven three-state machine with built-in pacing control. (b) \textbf{Architecture of SToP}: A lightweight pacing module for token generation rate regulation.}
    \label{fig:FDTC_SToP}
\end{figure}

FDTC continuously triggers state transitions through consecutive video frame inputs. 
At each frame, it dynamically determines whether to continue the current utterance, trigger a new response, or remain silent, based on the evolving visual context. The following details the three core states and their transition paths (Fig.~\ref{fig:model_architecture} and Fig.~\ref{fig:FDTC_SToP}(a)):

\textbf{Continuing State} is the most critical state. All triggered utterances enter this state (T → C) and remain looping within it until the utterance is completed. Upon receiving a frame in this state:
\textbf{(1) C → C:} If the current frame, when added to the context, still allows for the fluent decoding of the existing utterance buffer (indicating it belongs to the same semantic clip and can share the same utterance prefix), the controller decodes $m$ new tokens and remains in Continuing State.
\textbf{(2) C → S:} Similar to above, except decoding produces an EOS token (e.g., <|im\_end|>), indicating no further decoding is needed for the current utterance. It then transitions to Silent State.
\textbf{(3) C → T:} If appending the current frame to the context makes fluent decoding of the existing utterance buffer infeasible (suggesting a desire to trigger new content), FDTC triggers the beginning $m$ tokens of a new utterance and transitions to Triggered State.

\textbf{Triggered State} is entered immediately when a new utterance is triggered from any other state (C → T, S → T). Upon receiving a frame in this state:
\textbf{T → C:} 
The controller defaults to transitioning to the Continuing State to generate the remainder of the utterance. A direct T → T transition is undefined, as the initial $m$ tokens contain limited information and are prone to inheritance or prefix sharing with subsequent content, preventing distinct new events. Moreover, no T → T transitions were observed in our experiments.

\textbf{Silent State} is entered only after a complete utterance is generated in Continuing State (C → S), signifying the end of generation. FDTC remains silent, awaiting the next trigger. Upon receiving a frame in this state: 
\textbf{(1) S → S:} Similar to C → C, if appending the current frame to the context still allows fluent decoding of the existing (complete) utterance buffer (indicating the frame pertains to the same concluded event), FDTC does not trigger a new utterance and remains in Silent State.
\textbf{(2) S → T:} Similar to C → T, if decoding the utterance buffer fails after adding the frame (suggesting a desire to trigger new content), FDTC triggers the first $m$ tokens of a new response and transitions to Triggered State.

\subsection{Streaming Token Pacer}
Achieving human-like fine-grained synchrony requires the model to adapt its speaking rate to the pacing of visual events. Real-world video streams exhibit varying semantic densities, necessitating distinct narration tempos. For instance, a high-octane live sports scenario requires rapid-fire commentary (high tokens/sec) to keep up with the action, whereas a slow-paced cinematic shot demands sparse, relaxed narration (low tokens/sec). Relying on a fixed generation parameter $m$ (as denoted in~\ref{sec:FDTC}) inevitably leads to desynchronization, either lagging behind fast-paced visual changes or causing unnecessary stuttering during static scenes. To address this, we propose the Streaming Token Pacer (\textbf{SToP}), an adaptive module that dynamically modulates the generation rate ($m_i$ tokens per frame $t_i$) based on visual semantic density and real-time latency constraints.

\noindent \textbf{Content-Aware Rate Prediction.}
The core of SToP is a lightweight pacing module $\Phi_{\text{pacer}}$, which predicts the optimal linguistic density for the current visual context. 
Determining the appropriate narration speed requires considering both unfolding visual cues (such as motion and scene cuts) and the preceding linguistic rhythm to maintain a natural narrative flow. 
Thus, $\Phi_{\text{pacer}}$ operates over a multimodal history sequence $C_{\text{hist}} = \{ \mathbf{v}_{i-N}, \mathbf{t}_{i-N}, \dots, \mathbf{v}_{i-1}, \mathbf{t}_{i-1}, \mathbf{v}_i \}$, where $\mathbf{v}_j \in \mathbb{R}^{D}$ is a frame-level visual embedding, and $\mathbf{t}_j \in \mathbb{R}^{D}$ is a learnable embedding for the discrete token count category at frame $t_j$.
We employ a temporal window of $N$ frames (e.g., $N=4$) to capture short-term variations across visual and pacing signals, allowing the pacer to adapt its prediction based on recent context. A [CLS] token is prepended to the sequence to aggregate global context.

\noindent \textbf{Pacing Training.}
The pacer $\Phi_{\text{pacer}}$ is a light Transformer~\cite{vaswani2017attention} encoder (Fig.~\ref{fig:FDTC_SToP}(b)) that uses self-attention to capture real-time pacing behavior within $C_{\text{hist}}$. It contains $L=2$ Transformer blocks, providing a favorable trade-off between accuracy and efficiency. The output [CLS] token $\mathbf{z}_{\text{CLS}} \in \mathbb{R}^{D}$ is fed into a two-layer MLP with Softmax to predict a distribution over $K$ token-count categories:
\begin{equation}
\label{eq:pacer_arch}
    \mathbf{p}_i = \text{Softmax}\!\left(\text{MLP}(\mathbf{z}_{\text{CLS}})\right),
\end{equation}
where $\mathbf{p}_i \in \mathbb{R}^{K}$ denotes the predicted probability distribution. During inference, the final token-count prediction, $\hat{m}_i$, is derived from this distribution by taking the most probable category, although using the expected value is also a valid alternative.
We train $\Phi_{\text{pacer}}$ on a large-scale corpus of time-aligned video–transcript pairs, denoted as $\mathcal{D}$. Each transcript provides word-level timestamps that are temporally aligned with the corresponding video stream. Based on these timestamps, we compute the ground-truth token count $m_{i,\text{gt}}$ per frame $t_i$. This integer count is then mapped to a ground-truth category label, $y_{i,\text{gt}} \in \{0, 1, ..., K-1\}$. The model is optimized by minimizing the Cross-Entropy loss, a standard choice for classification tasks, over the entire dataset. For each video-transcript pair $(V, \mathcal{T})$ in $\mathcal{D}$, the loss is formulated as:
\begin{equation}
\label{eq:pacer_loss}
    \mathcal{L}_{\text{pacer}} = \frac{1}{|\mathcal{D}|} \sum_{(V, \mathcal{T}) \in \mathcal{D}} \sum_i \text{CE}(\mathbf{p}_i, y_{i,\text{gt}}),
\end{equation}
with label smoothing applied to the target label $y_{i,\text{gt}}$ to preserve ordinal relationships among adjacent token-count categories. This training encourages the pacer to learn frame-level pacing patterns consistent with human narration.

\noindent \textbf{On the Choice of Supervision Signal.}
Our ground-truth pacing is derived from word-level timestamps of human transcripts, i.e., we treat the instantaneous speaking rate of a human narrator as a proxy for the ideal narration tempo. We deliberately make this choice for two reasons. First, the supervision is \emph{behaviorally grounded}: human narrators (e.g., sports commentators or audio describers) naturally accelerate their speech when visual events become dense and slow down during static intervals, so the timestamped transcript already encodes a visually-driven pacing signal at scale, without requiring any manual rate annotation. Second, it is \emph{abundant and automatically obtainable}, allowing SToP to be trained on large corpora (Live-WhisperX-526K) rather than a small set of hand-labeled clips. We acknowledge that speaking rate is only a proxy for, and not identical to, an optimal visual-event-density tempo: it inherits speaker-specific habits and ASR timestamp noise, and may diverge from the ``ideal'' pace in atypical narration styles. Crucially, SToP is not used as a hard constraint but as a soft, content-aware prior whose output is always reconciled with the real-time latency budget via the cutoff in Eq.~\ref{eq:latency_cutoff_v2}; consequently, moderate label noise in the proxy mainly affects \emph{how many} tokens are emitted per frame, not \emph{whether} synchrony is maintained. Replacing this weak supervision with human-annotated pacing, or with multi-narrator consensus targets, is a natural direction for future work.

\noindent \textbf{Latency-Constrained Adaptive Decoding.}
While $\hat{m}_i$ provides a content-aware token target, practical streaming synchrony requires strict real-time constraints. The total processing time for frame $t_i$, $T_{\text{total}}^{(i)}$, includes visual perception latency $T_{\text{perc}}$ and cumulative generation time $\sum_{j=1}^{k} T_{\text{gen}}^{(j)}$ for $k$ tokens. To ensure stutter-free playback without buffering, $T_{\text{total}}^{(i)}$ must not exceed the frame interval $\Delta t = 1/\text{FPS}$.

SToP enforces this budget via a hard latency cutoff during inference. 
As language decoding, the final number of emitted tokens, $m_i$, is capped by the more restrictive of two upper bounds: the pacer's content-aware prediction ($\hat{m}_i$) and the maximum number of tokens feasible within $\Delta t$: 
\begin{equation}
\label{eq:latency_cutoff_v2}
m_i = \min \left( 
    \hat{m}_i,\;
    \max \left\{ k \;\bigg|\;
        T_{\text{perc}} + \sum_{j=1}^{k} T_{\text{gen}}^{(j)} \le \Delta t
    \right\}
\right).
\end{equation}
During this process, generation also terminates immediately if the model emits an EOS token (e.g., <|im\_end|>). This dual-constraint mechanism preempts generation when the time budget $\Delta t$ is exhausted, prioritizing streaming synchronization over completing the content-aware target $\hat{m}_i$. 
By combining $\Phi_{\text{pacer}}$ with this strict hardware-aware latency constraint, SToP achieves adaptive and stutter-free video-language streaming across diverse real-world scenarios.

\section{Algorithm}
We present LyraV, a dynamic response–silence decoding model designed to determine optimal response timing for online video understanding. It introduces a decoding gate that selectively triggers caption generation based on a streaming verification mechanism, and incorporates a Streaming Token Pacer (SToP) module to adaptively regulate the number of tokens generated at each step.
At the core of LyraV is a Frame-Driven Transition Controller (FDTC) that governs the generation process through three distinct states:  (1) \textbf{Triggered State}: Activated when the verification perplexity exceeds a threshold, initiating new decoding. (2) \textbf{Continuing State}: Continues generation if the response is incomplete. (3) \textbf{Silent State}: Maintains silence when the response is complete and no new content is required.  

Formally, at each triggered decoding step \( t_i \), we compute the perplexity of the generated caption \([Rsp]\) as:  
\[
\text{PPL}^{t_i}([Rsp]) = \sqrt[N]{\frac{1}{P([Rsp] \mid [Ctx^{< t_i}], [Frm^{t_i}])}},
\]  
where \( N \) is the number of tokens in \([Rsp]\), and \( P(\cdot) \) is the autoregressive probability derived from token logits.
For each incoming frame \([Frm^{t_j}]\), LyraV performs a single forward pass to verify the latest caption’s validity by recomputing \( \text{PPL}^{t_j}([Rsp]) \).  
If \( \text{PPL}^{t_j}([Rsp]) > \alpha \cdot \text{PPL}^{t_i}([Rsp]) \) (where \( \alpha \) is a tunable scaling factor), the system enters the Triggered State. Here, the SToP module predicts the number of tokens \( m_j \) to decode based on the current frame and a temporal window \( W \). The model then generates the first \( m_j \) new tokens using the context \([Ctx^{< t_j}; Rsp; Frm^{t_j}]\), appends the frame–token pair to \( W \), moves the previous \([Rsp]\) to \([Ctx]\), updates \([Rsp]\) to the new tokens, and sets the checkpoint timestamp \( t_i \leftarrow t_j \).

If the perplexity condition is not met, the system checks whether the last token of \([Rsp]\) is an end-of-sequence (<EOS>) token.  
If not, it enters the Continuing State, where SToP again predicts \( m_j \), and generation continues for up to \( m_j \) new tokens (or until <EOS>) using \([Ctx^{< t_j}; Frm^{t_j}; Rsp]\). The new tokens are appended to \([Rsp]\), and \( W \) and \( t_i \) are updated accordingly.  
If <EOS> is already present, the system enters the Silent State, preserving the current response and context without further decoding.
When no context exists initially, LyraV performs an initial decoding using only the current frame, with \( m_j \) determined by SToP.
This lightweight verification mechanism, combined with token-level pacing, ensures adaptive response timing that balances accuracy and efficiency. Under the same model architecture, LyraV achieves faster inference than always decoding until EOS, while maintaining temporal coherence and minimizing latency in streaming video understanding. The detailed algorithm procedure is presented in pseudocode in Algorithm~\ref{alg:LyraStar}.

\begin{algorithm}
\footnotesize
\SetAlgoLined 
\DontPrintSemicolon
\SetKwInOut{KwData}{Initialize}
\SetKw{KwResult}{Note:}
\KwIn{Video frame stream $\{[Frm^{t}]\}_{t=1}^T$} 
\KwOut{Dynamically generated caption $[Rsp]$}
\KwData{$[Rsp],[Ctx] \leftarrow \emptyset$, checkpoint timestamp $t_i \leftarrow 0$, temporal window $W\leftarrow \emptyset$}
\For{each incoming frame $[Frm^{t_j}]$}{
  \eIf{$[Ctx] \neq \emptyset$}{
      Compute verification perplexity:\;
      $\text{PPL}^{t_j}([Rsp]) = \sqrt[N]{\frac{1}{P([Rsp] \mid [Ctx^{< t_j};Frm^{t_j}])}}$\;
      \eIf{$\text{PPL}^{t_j}([Rsp]) > \alpha \text{PPL}^{t_i}([Rsp])$}{
          \tcp{Triggered State}
          $m_j = \text{SToP}(W;Frm^{t_j})$ \tcp*{Predict the token number to decode}
          Activate new decoding: Generate the first $m_j$ new tokens using $[Ctx^{< t_j};Rsp;Frm^{t_j}]$\;
          Append $[Frm^{t_j};m_j]$ to $W$\;
          Append $[Rsp]$ to $[Ctx]$\;
          Update $[Rsp] \leftarrow \text{new tokens}$\;
          $t_i \leftarrow t_j$ %
          \tcp*{Update checkpoint timestamp}
      }
      {
          \eIf{$[Rsp][-1] \neq \text{<EOS>}$}{
            \tcp{Continuing State}
            $m_j = \text{SToP}(W;Frm^{t_j})$\;
            Continue generating $m_j$ new tokens until $\text{<EOS>}$ using $[Ctx^{< t_j};Frm^{t_j};Rsp]$\;
            Append $[Frm^{t_j};m_j]$ to $W$\;
            Append new tokens to $[Rsp]$\;
            $t_i \leftarrow t_j$ 
          }
          {
            \tcp{Silent State}
            Keep silent\; 
          }
      }
  }
  {
      $m_j = \text{SToP}(W;Frm^{t_j})$\;
      Perform initial decoding to generate the first $m_j$ new tokens using $[Frm^{t_j}]$\;
      Append $[Frm^{t_j};m_j]$ to $W$\;
      $[Rsp] \leftarrow \text{new tokens}$\;
  }
  Append $[Frm^{t_j}]$ to $[Ctx]$\;
}
Append $[Rsp]$ to $[Ctx]$\;
\KwResult{For clarity, $[Ctx]$ updates only explicitly; $[Frm]$ and $[Rsp]$ do not auto-update it.}
\caption{Frame-Driven Transition Controller (FDTC)}
\label{alg:LyraStar}
\end{algorithm}

\definecolor{highlightcolor}{HTML}{E6F0FF}

\begin{table*}[t]
  \centering
  \setlength{\tabcolsep}{4pt} 
  \renewcommand{\arraystretch}{0.60} 
  \caption{Online performance evaluation on real-time narration (Captioning) and streaming question answering (QA) tasks. $\uparrow$ indicates higher is better, and $\downarrow$ indicates lower is better. "-" indicates the inability to perform the required test or not reported.}
  \label{tab:online_exp}
  \footnotesize
  \resizebox{\textwidth}{!}{
  \begin{tabular}{@{}l cccc | ccc | cc@{}}
    \toprule
    \multirow{7}{*}{\textbf{Method}} & \multicolumn{7}{c}{\textbf{Real-time Captioning}} & \multicolumn{2}{c}{\textbf{Streaming QA}} \\
    \cmidrule(lr){2-8} \cmidrule(lr){9-10}
    & \multicolumn{4}{c}{\textbf{OmniStar-RNG (1fps)}} & \multicolumn{3}{c}{\textbf{Ego4D Narrations Stream}} & \makecell{\textbf{StreamingBench} \\ \textbf{(Real-Time)}} & \makecell{\textbf{OVO-Bench} \\ \textbf{(Overall)}} \\
    \cmidrule(lr){2-10}
    & \textbf{SS $\uparrow$} & \textbf{NF $\uparrow$} & \textbf{RL $\downarrow$} & \textbf{rFPS$\uparrow$} & \textbf{PPL$\downarrow$} & \textbf{TimeDiff$\downarrow$} & \textbf{TokAcc$\uparrow$} & \textbf{Avg. Score} & \textbf{Avg. Score} \\
    \midrule
    Human & 6.73 & 7.17 & 1.08 & - & - & - & - & 91.46 & 92.81 \\
    \midrule
    \multicolumn{10}{c}{\textit{Offline Models (Fixed decoding)}} \\
    \midrule
    Gemini 1.5 Pro~\cite{team2024gemini} & 5.34 & 5.69 & - & - & - & - & - & 75.69 & 63.00 \\
    GPT-4o~\cite{achiam2023gpt} & 5.03 & 5.46 & - & - & - & - & - & 73.28 & 59.54 \\
    LLaVA-Video~\cite{zhang2024llavanext-video} & 3.40 & 2.88 & - & - & - & - & - & 68.52 & 53.10 \\
    InternVideo2.5~\cite{wang2025internvideo2} & 4.32 & 3.61 & - & - & - & - & - & 72.68 & 53.21 \\
    MiniCPM-V2.6~\cite{yao2024minicpm} & 4.34 & 4.13 & - & - & - & - & - & 67.44 & 51.47 \\
    Qwen2.5-VL~\cite{bai2025qwen2} & 4.42 & 4.24 & - & - & - & - & - & 73.68 & 54.20 \\
    \midrule
    \multicolumn{10}{c}{\textit{Online Models (Proactive Response)}} \\
    \midrule
    VideoLLM-online~\cite{chen2024videollm} & 1.68 & 0.59 & 2.67 & 3.37 & 2.43 & 2.04 & 0.48 & 35.99 & 12.80 \\
    MMDuet~\cite{wang2024videollm} & 1.93 & 2.69 & 2.54 & 0.91 & 4.51 & 1.97 & 0.39 & 61.01 & 39.85 \\
    Dispider~\cite{qian2025dispider} & 2.51 & 3.24 & 2.88 & 1.26 & - & - & - & 67.63 & 41.78 \\
    LiveStar~\cite{yang2025livestar} & 3.19 & \textbf{4.25} & 1.91 & 3.82 & 1.97 & 1.76 & 0.61 & 71.92 & 50.34 \\
    \rowcolor{highlightcolor}
    \textbf{LyraV (Ours)} & \textbf{3.37} & 4.19 & \textbf{1.82} & \textbf{3.89 }& \textbf{1.94} & \textbf{1.69} & \textbf{0.62} & \textbf{72.78} & \textbf{50.97} \\
    \bottomrule
  \end{tabular}
  }
\end{table*}

\section{Experiments}
\subsection{Experimental Setup}
\textbf{Datasets.}
To comprehensively evaluate LyraV, we conduct experiments across 8 video-language tasks. For online settings, LyraV is assessed on real-time narration (OmniStar-RNG~\cite{yang2025livestar}, Ego4D Narration Stream~\cite{grauman2022ego4d}) and streaming QA (StreamingBench~\cite{lin2024streamingbench}, OVO-Bench~\cite{li2025ovo}, OVBench~\cite{huang2025online}), testing its coherence, temporal alignment, and real-time reasoning. Offline evaluation employs MVBench~\cite{li2024mvbench}, LongVideoBench~\cite{wu2024longvideobench}, and VideoMME~\cite{fu2025video} to measure general video understanding. Additionally, to address the lack of fine-grained synchrony metrics, we design a dedicated evaluation on OmniStar, quantifying synchrony rate and content coherence under different decoding strategies.

\noindent \textbf{Implementation Details.}
LyraV builds on a frozen, pre-trained online Video-LLM backbone (InternViT~\cite{chen2024expanding}+ InternLM2.5-7B~\cite{cai2024internlm2})~\cite{yang2025livestar}, keeping it frozen for video understanding and language generation. To enable streaming video-language synchronization, we introduce two modules: a Frame-Driven Transition Controller (FDTC), which is training-free and reuses a verification-based decoding signal but introduces a novel stateful, prefix-level continuation mechanism (the Continuing State) on top of it, and a lightweight Streaming Token Pacer (SToP), implemented as a two-layer Transformer encoder. SToP is trained on Live-WhisperX-526K to predict token-to-frame alignment using a 4-frame history, optimized with AdamW ($lr=1\times10^{-4}$, batch size=4) and Cross-Entropy loss, with input frames resized to 448×448, utilizing an NVIDIA H100 GPU.

\subsection{Evaluation Metrics}
This section provides detailed definitions and computational procedures for the evaluation metrics used in our online narration and synchrony experiments.

\subsubsection{Metrics for Online Experiments}
\textbf{Semantic Score (SS).} The Semantic Score evaluates the semantic alignment between the full model-generated narrative and the corresponding ground-truth narrative. We adopt GPT-4o as an automatic judge, following the now-standard ``LLM-as-a-judge'' protocol whose ratings have been repeatedly shown to correlate strongly with human judgments and to outperform conventional $n$-gram metrics in alignment with human preference~\cite{zheng2023judging,liu2023geval,chiang2023alternative}. For each video, we concatenate all generated utterances $\{u_1, u_2, \dots, u_k\}$ into a single narrative paragraph, $N_{\text{model}} = u_1 \circ u_2 \circ \dots \circ u_k,$ and compare it with the holistic ground-truth narrative $N_{\text{gt}}$. GPT-4o assigns a semantic score $S_{\text{SS}}(v)$ based on the following three dimensions:

\begin{itemize}
    \item \textbf{Content Accuracy} evaluates whether the generated narrative preserves factual correctness and semantic alignment with the ground truth. Narratives receiving a score near 10 demonstrate almost perfect correspondence to $N_{\text{gt}}$, retaining all essential meanings without distortion. Scores in the 7--9 range indicate minor deviations that do not compromise the core interpretation. Scores of 4--6 emerge when portions of the content are partially incorrect or missing key factual elements. Narratives scoring 1--3 diverge substantially, containing largely incorrect or irrelevant material, whereas a score of 0 reflects complete semantic mismatch or meaningless output.
    \item \textbf{Detail Relevance} concerns the precision and contextual grounding of descriptive details. A score near 10 is assigned when nearly all included details are specific, accurate, and directly supported by events in $N_{\text{gt}}$. Scores of 7--9 indicate mostly relevant details with slight redundancy or mild nonspecific phrasing. Scores of 4--6 correspond to narratives mixing relevant and irrelevant content, diluting informativeness. When the majority of provided details are generic or unrelated, the narrative receives 1--3 points. A score of 0 indicates entirely ungrounded or nonsensical descriptions.
    \item \textbf{Event Coverage} measures whether the generated narrative captures the essential events of the ground-truth storyline. A score of 10 reflects complete coverage of all major actions and states described in $N_{\text{gt}}$. Scores of 7--9 indicate that main events are preserved but minor dynamic details are missing. When significant elements of the narrative arc are omitted, the score falls into the 4--6 range. Narratives that retain only small fragments of the event structure receive 1--3 points, while a score of 0 indicates no meaningful coverage of the underlying events.
\end{itemize}

\textbf{Narrative Fluency (NF).}
Narrative Fluency evaluates the overall readability, continuity, and stylistic coherence of the full generated narrative. Using the same concatenated model narrative $N_{\text{model}}$, GPT-4o produces a fluency score $S_{\text{NF}}(v)$ based on three dimensions:

\begin{itemize}
    \item \textbf{Logical Transitions} assess whether ideas progress smoothly with clear causal, temporal, or inferential connections. Narratives scoring near 10 exhibit consistently coherent and natural transitions. Scores of 7--9 maintain overall clarity but contain minor jumps or mildly awkward reasoning. Scores of 4--6 reflect noticeable disruptions in logic that may require substantial reader inference to follow. A score of 1--3 indicates frequent contradictions or disjoint reasoning, while 0 reflects an absence of coherent logical structure.
    \item \textbf{Temporal Flow} evaluates whether events in the concatenated narrative unfold in a stable and interpretable temporal order. Scores close to 10 correspond to a fluent storytelling structure with accurate temporal anchoring across segments. Scores of 7--9 remain largely chronological but contain slight inconsistencies. Narratives scoring 4--6 display confusing or partially reversed event sequences, reducing interpretability. Scores of 1--3 reflect highly inconsistent temporal ordering, and 0 corresponds to a wholly incoherent or unintelligible temporal structure.
    \item \textbf{Stylistic Consistency} concerns the stability of tone, diction, and linguistic form across the combined narrative. A score of 10 reflects a uniform and fluent writing style with no abrupt shifts. Scores of 7--9 show small stylistic variations that do not affect readability. When multiple inconsistent stylistic choices appear, the narrative receives 4--6 points. Scores of 1--3 indicate substantial stylistic instability—such as rapid tone shifts or uneven phrasing—while 0 corresponds to stylistically incoherent or unreadable text.
\end{itemize}

\textbf{Response Latency (RL).}
This metric quantifies the cumulative temporal deviation between model-generated responses and the ground-truth event timeline. It penalizes delayed outputs, missed events, and redundant responses under a unified formulation. For our model, \textit{LyraV}, the response time $t_{\text{resp}}$ is defined as the moment its FDTC enters the \textbf{Triggered state}, which differs from conventional online models where $t_{\text{resp}}$ corresponds to the instant an explicit response is generated.

We treat the ground-truth narrative $N_{\text{gt}}$ as a sequence of temporally annotated events $\{e_1, e_2, \dots, e_N\}$. For each event $e_i$ spanning the interval $[t_{\text{start},i}, t_{\text{end},i}]$, we collect all model responses whose timestamps fall within this window:
\[
M_i = \{\, r_j \mid t_{\text{resp},j} \in [t_{\text{start},i},\, t_{\text{end},i}] \,\}.
\]
The penalty for event $e_i$, denoted $L(e_i)$, equals the full event duration if no model response is present ($|M_i| = 0$). Otherwise, it is the cumulative latency of all responses in $M_i$, where each response contributes $t_{\text{resp},j} - t_{\text{start},i}$. The final RL score is:
\[
\begin{split}
\text{RL} = \frac{1}{N} \sum_{i=1}^N \Big(
& \mathbb{I}[|M_i| = 0] \cdot (t_{\text{end},i} - t_{\text{start},i}) \\
+ \;& \mathbb{I}[|M_i| > 0] \cdot 
\sum_{r_j \in M_i} \big(t_{\text{resp},j} - t_{\text{start},i}\big)
\Big),
\end{split}
\]
where $N$ is the total number of ground-truth events and $\mathbb{I}[\cdot]$ is the indicator function. This formulation naturally assigns larger penalties to late responses, full-duration penalties to missed events, and additional penalties to redundant responses since all responses in $M_i$ are cumulatively included.

\textbf{Real-time FPS (rFPS).}
We report an end-to-end throughput metric that reflects the system's effective output processing speed under online, actively responding conditions. Specifically, the measured duration $T_{\mathrm{total}}$ corresponds to the actual wall-clock time from the moment the \emph{first video frame} enters the streaming pipeline to the moment the \emph{final token} of the model's last response is produced. Let $N_{\mathrm{frames}}$ denote the total number of frames processed under this setting. The real-time frames-per-second is then defined as
\[
\text{rFPS} \;=\; \frac{N_{\mathrm{frames}}}{T_{\mathrm{total}}}.
\]
This definition follows the conventional notion of throughput-based FPS, but we denote it as rFPS to distinguish it from the Sync Rate (SR), which instead relies on the \emph{input playback} frame rate to determine a per-frame time budget rather than measuring system-level output efficiency.

\begin{table*}[t!]
\centering
\caption{\textbf{Evaluation results on OVO-Bench.} OVO-Bench comprises three challenging categories: (i) \textit{Real-Time Visual Perception}, (ii) \textit{Backward Tracing}, and (iii) \textit{Forward Active Responding}.}
\label{tab:ovo_bench_results}
\resizebox{\textwidth}{!}{
\begin{tabular}{l c | cccccc c | cccc | cccc | c}
\toprule
\multirow{2}{*}{\textbf{Model}} & \multirow{2}{*}{\textbf{\#Frames}} & \multicolumn{7}{c|}{\textbf{Real-Time Visual Perception}} & \multicolumn{4}{c|}{\textbf{Backward Tracing}} & \multicolumn{4}{c|}{\textbf{Forward Active Responding}} & \multicolumn{1}{c}{\textbf{Overall}} \\
\cmidrule(lr){3-9} \cmidrule(lr){10-13} \cmidrule(lr){14-17} \cmidrule(lr){18-18}
& & \multicolumn{1}{c}{OCR} & \multicolumn{1}{c}{ACR} & \multicolumn{1}{c}{ATR} & \multicolumn{1}{c}{STU} & \multicolumn{1}{c}{FPD} & \multicolumn{1}{c}{OJR} & \multicolumn{1}{c|}{Avg.} & \multicolumn{1}{c}{EPM} & \multicolumn{1}{c}{ASI} & \multicolumn{1}{c}{HLD} & \multicolumn{1}{c|}{Avg.} & \multicolumn{1}{c}{REC} & \multicolumn{1}{c}{SSR} & \multicolumn{1}{c}{CRR} & \multicolumn{1}{c|}{Avg.} & \multicolumn{1}{c}{Avg.} \\
\midrule
Human & - & 94.0 & 92.6 & 94.8 & 92.7 & 91.1 & 94.0 & 93.2 & 92.6 & 93.0 & 91.4 & 92.3 & 95.5 & 89.7 & 93.6 & 92.9 & 92.8 \\
\midrule
\multicolumn{18}{c}{\textit{Offline Models}} \\
\midrule
Gemini 1.5 Pro & 1fps & 85.91 & 66.97 & 79.31 & 58.43 & 63.37 & 61.96 & 69.32 & 58.59 & 76.35 & 52.64 & 62.54 & 35.53 & 74.24 & 61.67 & 57.15 & 63.00 \\
GPT-4o & 64 & 69.80 & 64.22 & 71.55 & 51.12 & 70.30 & 59.78 & 64.46 & 57.91 & 75.68 & 48.66 & 60.75 & 27.58 & 73.21 & 59.40 & 53.40 & 59.54 \\
LLaVA-Video & 64 & 69.80 & 59.63 & 66.38 & 50.56 & 72.28 & 61.41 & 63.34  & 51.18 & 64.19 & 9.68 & 41.68 & 34.10 & 67.57 & 60.83 & 54.17 & 53.06 \\
LLaVA-OneVision & 64 & 67.11 & 58.72 & 69.83 & 49.45 & 71.29 & 60.33 & 62.79 & 52.53 & 58.78 & 23.66 & 44.99 & 24.79 & 66.93 & 60.83 & 50.85 & 52.88 \\
Qwen2-VL & 64 & 69.13 & 53.21 & 63.79 & 50.56 & 66.34 & 60.87 & 60.65 & 44.44 & 66.89 & 34.41 & 48.58 & 30.09 & 65.66 & 50.83 & 48.86 & 52.70 \\
InternVL-V2 & 64 & 68.46 & 58.72 & 68.97 & 44.94 & 67.33 & 55.98 & 60.73 & 43.10 & 61.49 & 27.41 & 44.00 & 25.79 & 57.55 & 52.92 & 45.42 & 50.05 \\
LongVU & 1fps & 55.70 & 49.54 & 59.48 & 48.31 & 68.32 & 63.04 & 57.40 & 43.10 & 66.22 & 9.14 & 39.49 & 16.62 & 69.00 & 60.0 & 48.54 & 48.48 \\
\midrule
\multicolumn{18}{c}{\textit{Online Models (Proactive Response)}} \\
\midrule
VideoLLM-online & 2fps & 8.05 & 23.85 & 12.07 & 14.04 & 45.54 & 21.20 & 20.79 & 22.22 & 18.80 & \textbf{12.18} & 17.73 & - & - & - & - & - \\
Dispider & 1fps & 57.72 & 49.54 & 62.07 & 44.94 & 61.39 & 51.63 & 54.55 & 48.48 & 55.41 & 4.30 & 36.06 & 18.05 & 37.36 & 48.75 & 34.72 & 41.78 \\
LiveStar~\cite{yang2025livestar} & 1fps & \textbf{74.62} & 65.70 & \textbf{67.40} & 50.20 & 68.10 & 62.06 & 64.68 & \textbf{56.94} & 67.30 & 9.89 & 44.71 & 31.50 & 43.70 & 49.69 & 41.63 & 50.34 \\
\rowcolor{highlightcolor}
\textbf{LyraV (Ours)} & 1fps & 74.45 & \textbf{65.93} & 67.13 & \textbf{50.28} & \textbf{68.21} & \textbf{62.34} & \textbf{64.72} & 56.41 & \textbf{67.85} & 10.01 & \textbf{44.76} & \textbf{33.11}  & \textbf{45.12} & \textbf{52.03} & \textbf{43.42} & \textbf{50.97} \\
\bottomrule
\end{tabular}
}
\end{table*}

\subsubsection{Synchrony Evaluation Metrics}

\textbf{Sync Rate (SR).}
This metric evaluates how well a model adheres to a predefined playback rhythm during continuous video streaming. For a video played at frame rate $FPS_{\text{video}}$, the per-frame time budget is $\Delta t_{\text{budget}} = 1 / FPS_{\text{video}}$. For each frame $i$, let $t_{\text{proc}}^{(i)}$ denote the actual processing time; we define the effective processing time as $t_{\text{eff}}^{(i)} = \max(t_{\text{proc}}^{(i)}, \Delta t_{\text{budget}})$ so that delays exceeding the frame budget are fully penalized while on-time completions are treated as meeting the budget exactly. The Sync Rate is then the ratio between the ideal total processing time and the accumulated effective processing time:
\[
\text{SR} = 
\frac{N_{\text{frames}} \cdot \Delta t_{\text{budget}}}
     {\sum_{i=1}^{N_{\text{frames}}} t_{\text{eff}}^{(i)}} \times 100\%.
\]
An SR of $100\%$ indicates perfect real-time synchrony, whereas lower values directly reflect cumulative perceptual delays.

\subsection{Online Experiments}
We conduct a comprehensive online evaluation of LyraV under two core streaming video tasks: streaming captioning and QA. For captioning, all models generate outputs under their native proactive response mechanisms, evaluating both response timing and content quality. For streaming QA, we strictly follow official benchmark settings~\cite{11168273,11097075}.

\noindent\textbf{Evaluation Metrics.}
We employ tailored online metrics. For real-time captioning on OmniStar, we assess across narrative quality, temporal precision, and processing efficiency.
\begin{itemize}
    \item \textbf{Semantic Score (SS)} evaluates semantic alignment of the generated narrative with ground truth via GPT-4o (0–10), considering accuracy, relevance, and event coverage.
    \item \textbf{Narrative Fluency (NF)} measures the overall coherence of the concatenated narrative, focusing on logical transitions, temporal flow, and stylistic consistency.
    \item \textbf{Response Latency (RL)} quantifies the temporal deviation between responses and GT event timings, where both missed and redundant responses are penalized. 
    \item \textbf{Real-time FPS (rFPS)} computes end-to-end video processing speed (\textit{output}) in frames per second, indicating how efficiently it handles streaming input (distinct from the FPS (\textit{input}) definition in SR, detailed in Sec. \ref{sec:SVLS_experiments}). 
\end{itemize}

\noindent For the Ego4D Narration Stream, we follow the evaluation metrics used in prior works, including \textbf{Perplexity}, \textbf{TimeDiff} and \textbf{Token Accuracy (TokAcc)}. For StreamingBench and OVO-Bench, we adopt their officially defined accuracy metrics. 

\noindent\textbf{Results.}
As shown in Tab.~\ref{tab:online_exp}, LyraV's gains concentrate on the time-sensitive captioning setting: on OmniStar-RNG it improves semantic score by 5.64\% and reduces response latency by 4.95\% over its backbone LiveStar, where response timing and synchrony are explicitly rewarded. On OmniStar-RNG it slightly trails LiveStar in Narrative Fluency due to FDTC-induced incomplete responses, yet this trade-off enables richer multi-frame reasoning. By contrast, on the streaming QA benchmarks (StreamingBench, OVO-Bench) LyraV achieves 72.78 vs.\ LiveStar's 71.92 and 50.97 vs.\ LiveStar's 50.34. These marginal gains (1.19\% and 1.25\%) reflect the effect of our online decoding strategy in stabilizing responses across frame-by-frame context shifts, suggesting that the Continuing State mechanism provides concrete benefits in time-sensitive QA reasoning. This is expected rather than a shortcoming: these QA benchmarks score static, short-answer accuracy and do not measure real-time responsiveness or speak-while-watching synchrony, so a synchrony-oriented control layer wrapped around a frozen backbone is, by design, expected to \emph{preserve} rather than raise such scores. The substantive advantages of LyraV instead surface on the synchrony protocol (Tab.~\ref{tab:synchrony_exp}). We qualitatively attribute the preserved QA accuracy to an incremental refinement behavior observed during inference, illustrated in Sec.~\ref{subsec:insight}.

\subsubsection{Per-Task Results on OVO-Bench}
The rightmost section of Table~\ref{tab:online_exp} also includes the average evaluation outcomes on real-time tasks from OVO-Bench~\cite{li2025ovo}. The corresponding metric-specific analyses for this benchmark are illustrated separately in Table~\ref{tab:ovo_bench_results}.
In the realm of online models with proactive response capabilities, LyraV reaches an overall average score of 50.97, modestly ahead of its frozen backbone LiveStar (50.34; the gap is still small but consistent) and clearly above other online models such as VideoLLM-online (20.79, averaged based on available results) and Dispider (41.78). We note that OVO-Bench scores question-answering accuracy rather than real-time synchrony, so this slight improvement reflects our control layer’s ability to stabilize temporal reasoning rather than a fundamental increase in static perception capability. In critical sub-tasks like OCR (74.45) and ACR (65.93), LyraV competes closely with several offline models, including LLaVA-Video and LLaVA-OneVision, despite operating under real-time constraints.
Given that LyraV's only lower metric compared to VideoLLM-online is HLD, its hallucination detection capability still requires improvement.
When compared to offline models, which often benefit from full video access, LyraV holds its own remarkably well. While offline counterparts such as Gemini 1.5 Pro (63.00) and GPT-4o (59.54) achieve higher overall scores, LyraV's performance remains highly competitive, especially considering its online operation and reliance on incremental, streaming comprehension.

\subsubsection{Per-Task Results on OVBench}
Based on the comprehensive evaluation conducted with OVBench (as shown in Table~\ref{tab:ovbench_results}), our LyraV model is competitive among open-source online video MLLMs, attaining an average score of 46.8. This represents a clear improvement over other online methods such as MovieChat (30.9) and Flash-Vstream (31.2). Notably, LyraV remains competitive with several open-source offline video MLLMs (including InternVL2-7B (48.7) and LLaVA-Onevision (49.5)) and narrows the performance gap with proprietary models like Gemini-1.5-Flash (50.7).
LyraV performs well across multiple tasks, achieving leading results among online models on key subtasks such as OP (68.9 under THV), PR (65.0 under PM), and AT (69.2 under STP). These outcomes suggest its robustness in temporal reasoning and proactive response, and indicate the model's effectiveness in real-time video understanding.
However, LyraV trails behind the reactive response model Flash-Vstream in metrics such as Action Persistence (AP) and Object Existence State (OES). This indicates that proactive response models still have room for improvement in addressing temporal hallucination and enhancing verification performance.

\begin{table*}[t!]
\centering
\caption{\textbf{Results on OVBench.} Among open-source online video MLLMs, LyraV attains the strongest average score on OVBench, improving over prior online methods and narrowing the gap with leading offline and proprietary models.}
\label{tab:ovbench_results}
\resizebox{\textwidth}{!}{
\begin{tabular}{l c | ccc ccc ccc cc cc ccc | c}
\toprule
\multirow{2}{*}{\textbf{Task Name}} & \multirow{2}{*}{\textbf{Size}} & \multicolumn{3}{c}{\textbf{FP}} & \multicolumn{3}{c}{\textbf{THV}} & \multicolumn{3}{c}{\textbf{PM}} & \multicolumn{2}{c}{\textbf{SP}} & \multicolumn{2}{c}{\textbf{STP}} & \multicolumn{3}{c|}{\textbf{TP}} & \multirow{2}{*}{\textbf{AVG}} \\
\cmidrule(lr){3-5} \cmidrule(lr){6-8} \cmidrule(lr){9-11} \cmidrule(lr){12-13} \cmidrule(lr){14-15} \cmidrule(lr){16-18}
\textbf{Subset Name} & & AA & GSP & MP & AP & SV & OP & AR & PR & TR & AL & OP & AT & OT & AS & SL & OES & \\
\midrule
\multicolumn{19}{c}{\textit{Proprietary MLLMs}} \\
\midrule
Gemini-1.5-Flash~\cite{team2024gemini} & - & 71.4 & 53.6 & 21.9 & 56.5 & 60.8 & 40.6 & 36.7 & 47.9 & 62.5 & 32.3 & 37.5 & 87.0 & 50.0 & 83.3 & 22.3 & 46.9 & 50.7 \\
\midrule
\multicolumn{19}{c}{\textit{Open-source Offline Video MLLMs}} \\
\midrule
InternVL2~\cite{chen2024far} & 7B & 52.6 & 60.2 & 27.6 & 57.5 & 52.0 & 58.5 & 38.8 & 67.1 & 58.3 & 38.1 & 31.3 & 87.4 & 37.0 & 75.4 & 31.4 & 5.9 & 48.7 \\
InternVL2~\cite{chen2024far} & 4B & 57.7 & 57.0 & 14.4 & 59.2 & 49.4 & 60.0 & 30.3 & 61.8 & 46.3 & 30.9 & 20.1 & 83.0 & 32.3 & 70.7 & 29.4 & 3.4 & 44.1 \\
LLaMA-VID~\cite{li2024llama} & 7B & 43.6 & 50.9 & 19.6 & 64.0 & 47.5 & 46.8 & 29.4 & 48.9 & 51.2 & 31.9 & 11.2 & 75.7 & 24.8 & 59.1 & 26.0 & 40.0 & 41.9 \\
LLaVA-Onevision~\cite{li2024llava} & 7B & 68.0 & 62.7 & 35.9 & 58.4 & 50.3 & 46.5 & 29.4 & 60.7 & 58.0 & 43.1 & 14.2 & 86.5 & 49.7 & 70.7 & 28.1 & 30.2 & 49.5 \\
LongVA~\cite{zhang2024long} & 7B & 64.1 & 56.5 & 29.5 & 54.9 & 51.9 & 34.8 & 35.3 & 55.6 & 57.7 & 31.6 & 3.4 & 67.4 & 44.7 & 80.0 & 26.7 & 4.0 & 43.6 \\
MiniCPM-V2.6~\cite{yao2024minicpm} & 7B & 33.3 & 35.9 & 15.0 & 59.2 & 50.8 & 55.1 & 25.0 & 37.4 & 41.7 & 26.6 & 11.8 & 98.3 & 36.3 & 66.1 & 26.4 & 6.2 & 39.1 \\
Qwen2-VL~\cite{Qwen2VL} & 7B & 60.3 & 66.1 & 22.1 & 54.9 & 51.5 & 51.1 & 37.8 & 64.4 & 69.3 & 35.3 & 28.5 & 97.0 & 49.4 & 65.1 & 30.8 & 11.7 & 49.7 \\
LITA~\cite{huang2024lita} & 7B & 19.2 & 24.5 & 19.9 & 40.8 & 48.9 & 24.9 & 3.1 & 27.3 & 6.4 & 6.9 & 14.6 & 35.2 & 23.9 & 27.4 & 0.5 & 3.4 & 20.4 \\
TimeChat~\cite{ren2024timechat} & 7B & 7.7 & 15.3 & 18.7 & 20.6 & 15.7 & 11.7 & 9.1 & 14.7 & 9.8 & 7.5 & 19.5 & 13.9 & 10.3 & 9.3 & 10.1 & 10.8 & 12.8 \\
VTimeLLM~\cite{huang2024vtimellm} & 7B & 37.2 & 23.4 & 15.0 & 64.8 & 43.8 & 53.2 & 25.9 & 38.8 & 32.5 & 25.9 & 20.4 & 40.9 & 6.8 & 48.4 & 43.5 & 8.6 & 33.1 \\
\midrule
\multicolumn{19}{c}{\textit{Open-source Online Video MLLMs (Proactive / Reactive Response)}} \\
\midrule
VideoLLM-Online~\cite{chen2024videollm} & 7B & 0.0 & 1.8 & 20.9 & 5.2 & 5.9 & 32.6 & 0.0 & 2.3 & 26.7 & 0.6 & 26.6 & 0.9 & 19.9 & 0.9 & 1.7 & 8.3 & 9.6 \\
MovieChat~\cite{song2024moviechat} & 7B & 23.1 & 27.5 & 23.6 & 58.4 & 43.9 & 40.3 & 25.6 & 31.1 & 23.9 & 26.9 & 39.6 & 24.4 & 28.9 & 29.3 & 25.5 & 21.9 & 30.9 \\
Flash-Vstream~\cite{zhang2024flash} & 7B & 26.9 & 37.6 & 23.9 & \textbf{60.1} & 41.9 & 40.0 & 23.4 & 35.3 & 26.1 & 24.7 & 28.8 & 27.0 & 21.4 & 29.8 & 25.6 & \textbf{26.8} & 31.2 \\
LiveStar~\cite{yang2025livestar} & 7B & 35.8 & \textbf{57.4} & 23.8 & 46.6 & 49.3 & \textbf{69.1} & 33.4 & 63.0 & 50.8 & 33.2 & 39.9 & \textbf{69.5} & 44.8 & \textbf{65.3} & 28.3 & 21.2 & 45.7 \\
\rowcolor{highlightcolor}
\textbf{LyraV (Ours)} & 7B & \textbf{37.2} & 57.1 & \textbf{25.0} & 48.1 & \textbf{50.8} & 68.9 & \textbf{34.7} & \textbf{65.0} & \textbf{52.3} & \textbf{34.6} & \textbf{41.4} & 69.2 & \textbf{46.6} & 65.0 & \textbf{29.5} & 22.8 & \textbf{46.8} \\
\bottomrule
\end{tabular}
}
\end{table*}

\subsection{Offline Experiments}
\begin{table}[t]
  \centering
  \caption{Offline performance evaluation on established video understanding benchmarks. All scores are average accuracy (\%).}
  \label{tab:offline_exp}

  \scalebox{1.00}{
  
  \begin{tabular}{@{} l c c cc @{}}
    \toprule
    \multirow{2}{*}{\textbf{Method}} & \multirow{2}{*}{\textbf{MVBench}} & \multirow{2}{*}{\makecell{\textbf{LongVideo-} \\ \textbf{Bench}}} & \multicolumn{2}{c}{\textbf{VideoMME}} \\
    \cmidrule(lr){4-5}
    & & & \makecell{w/o sub.} & \makecell{w/ sub.} \\
    \midrule
    VideoLLM-online & 33.90 & 24.20 & 26.90 & 29.90 \\
    MMDuet & 65.30 & 53.30 & 57.70 & \textbf{67.10} \\
    Dispider & 55.65 & 51.80 & 57.20 & - \\
    LiveStar & 66.95 & \textbf{57.00} & 60.80 & 64.40 \\
    \rowcolor{highlightcolor}
    \textbf{LyraV (Ours)} & \textbf{67.10} & 56.80 & \textbf{60.90}& 64.10\\
    \bottomrule
  \end{tabular}
}
\end{table}
To ensure our synchrony framework preserves fundamental video understanding, we evaluate LyraV on established offline benchmarks: MVBench~\cite{li2024mvbench}, LongVideoBench~\cite{wu2024longvideobench}, and Video-MME~\cite{fu2025video}. In this setting, models are evaluated offline with full-video access, isolating understanding and reasoning from real-time constraints.

\noindent\textbf{Results.}
Our offline results (Tab.~\ref{tab:offline_exp}) show that integrating our hierarchical control framework maintains the backbone model's performance without notable degradation. This is by design: in offline settings, our synchrony-oriented modules (FDTC and SToP) remain inactive, preserving the backbone's inference integrity. We also observe that our backbone compares favorably with online-focused models (e.g., VideoLLM-online, Dispider) on these offline benchmarks, which suggests that some real-time models may trade off general understanding for responsiveness. Overall, these results indicate that LyraV improves streaming synchrony while retaining the backbone's general video-language capabilities, rather than improving static accuracy.

\begin{table*}[t]
\centering
\setlength{\tabcolsep}{4pt}
\renewcommand{\arraystretch}{0.8}

\begin{minipage}[t]{0.52\textwidth}
\centering

\caption{Video-Language Synchrony Benchmark. The Un-truncated setting reflects real-world performance with full decoding (2fps input, real fps output), while Truncated setting evaluates performance under zero-latency constraints (2fps input \& output).}
\label{tab:synchrony_exp}
\scalebox{1.00}{
\small
\begin{tabular}{llccc}
\toprule
\textbf{Setting} & \textbf{Model} & \textbf{SR} $\uparrow$ & \textbf{NF} $\uparrow$ & \textbf{SS} $\uparrow$ \\
\midrule
\multirow{5}{*}{\makecell[l]{Un-truncated \\ (Full Decode)}} 
 & VideoLLM-online & 88.26\% & 0.63 & 1.95 \\
 & MMDuet & 56.81\% & 2.77 & 2.46 \\
 & Dispider & 69.05\% & 3.20 & 2.96 \\
 & LiveStar & 78.93\% & \textbf{4.19} & 3.44 \\
 & LiveCC & 92.41\% & 2.71 & 2.58 \\
 & \cellcolor{highlightcolor}\textbf{LyraV (Ours)} & \cellcolor{highlightcolor}\textbf{98.29\%} & \cellcolor{highlightcolor}4.07 & \cellcolor{highlightcolor}\textbf{3.62} \\
\midrule
\multirow{5}{*}{\makecell[l]{Truncated \\ (Zero-latency)}} 
 & VideoLLM-online & 100\% & 0.53 & 1.71 \\
 & MMDuet & 100\% & 1.91 & 1.26 \\
 & Dispider & 100\% & 2.99 & 2.75\\
 & LiveStar & 100\% & 3.95 & 3.38 \\
 & LiveCC & 100\% & 2.59 & 2.44 \\
 & \cellcolor{highlightcolor}\textbf{LyraV (Ours)} & \cellcolor{highlightcolor}100\% & \cellcolor{highlightcolor}\textbf{4.03} & \cellcolor{highlightcolor}\textbf{3.63} \\
\bottomrule
\end{tabular}
}
\end{minipage}
\hfill
\begin{minipage}[t]{0.46\textwidth}
\centering

\captionsetup{skip=6pt}
\caption{Ablation of control modules. Removing FDTC causes pronounced drop, while removing SToP leads to slight declines.}
\label{tab:ablation_control}
\small
\scalebox{1.00}{
\begin{tabular}{lccc}
\toprule
\textbf{Model Variant} & \textbf{SS} $\uparrow$ & \textbf{NF} $\uparrow$ & \textbf{RL} $\downarrow$ \\
\midrule
LiveStar (Backbone) & 3.19 & \textbf{4.25} & 1.91 \\
LyraV w/o FDTC & 2.21 & 1.87 & 2.10 \\
LyraV w/o SToP & 3.08 & 3.86 & 1.83 \\
\rowcolor{highlightcolor}
\textbf{LyraV (Full)} & \textbf{3.37} & 4.19 & \textbf{1.82} \\
\bottomrule
\end{tabular}
}
\vspace{1em} 

\captionsetup{skip=6pt}
\caption{Comparison of different Pacer architectures. Transformer Pacer achieves the best overall performance.}
\label{tab:pacer_comparison}
\small
\scalebox{1.00}{
\begin{tabular}{lccc}
\toprule
\textbf{Pacer Arch.} & \textbf{Params.} & \textbf{rFPS} $\uparrow$ & \textbf{Acc.} $\uparrow$ \\
\midrule
RNN Pacer & 6.32M & 187.53 & 78.14\% \\
LSTM Pacer & 12.63M & 114.06 & 82.97\% \\
\rowcolor{highlightcolor}
\textbf{Transformer Pacer} & 8.55M & \textbf{206.13} & \textbf{91.55\%} \\
\bottomrule
\end{tabular}
}
\end{minipage}

\end{table*}

\subsection{Video-Language Synchrony Experiments}
\label{sec:SVLS_experiments}
We conduct fine-grained video-language synchrony evaluation central to our work, to directly measure model performance under a strict "speak-while-watching" paradigm, a capability not assessed by existing benchmarks.

\noindent\textbf{Evaluation Protocol and Metrics.}
We propose two complementary settings that probe different aspects of synchrony. (1) \textbf{Realistic Truncated Setting}: 
Models process video at 2fps and must complete both perception and generation within each 0.5s frame budget; responses exceeding this budget are truncated. This tests incremental generation and coherence under hard real-time constraints.
(2) \textbf{Blocking Un-truncated Setting}: 
Video frames arrive at 2fps, but the model blocks incoming frames until the current response is fully decoded. This reveals the perceptual delays inherent in conventional "pause-and-decode" approaches.

We introduce a specialized metric to evaluate synchrony:
\begin{itemize}
    \item \textbf{Sync Rate (SR):} 
    Ratio of the model’s effective processing rate to the video playback rate. Latencies beyond the 0.5s budget are penalized. Higher SR (up to 100\%) indicates better synchrony and fewer perceptual delays.
\end{itemize}
We emphasize that SR alone is \emph{not} a sufficient measure of model quality: any model that simply truncates generation to fit the per-frame budget can trivially reach a high SR, and indeed all models attain 100\% SR once truncation is enforced. SR therefore primarily reflects \emph{whether} a model keeps pace with playback, not \emph{how well} it narrates. For this reason, we always report SR jointly with content-quality metrics (SS and NF), and treat the truncated, equal-budget setting—where SR is held fixed at 100\% for all models—as the fair comparison that isolates the contribution of FDTC and SToP from the mere act of truncation.

\noindent\textbf{Results.}
We read Tab.~\ref{tab:synchrony_exp} primarily through the truncated, equal-budget setting, where SR is fixed at 100\% for every model and the comparison thus reduces to content quality under identical real-time constraints. Here LyraV attains the best SS and NF, improving NF by 2.0\% and SS by 7.4\% over LiveStar; competing models struggle to produce complete content within the budget and frequently restart sentences after truncation, yielding fragmented narration. This indicates that LyraV's advantage stems from how FDTC and SToP schedule generation under a fixed budget, rather than from truncation alone. In the complementary untruncated setting, the SR gap (LyraV 98.29\% vs.\ LiveStar 78.93\%) largely reflects whether a model pauses perception to decode: pause-and-decode baselines accumulate perceptual delay, whereas LyraV's incremental emission keeps it near real time while retaining the best overall balance of synchrony and semantic quality.

\subsection{Ablation Study}
\textbf{Impact of Control Modules.}
We compare the full LyraV with three variants: (1) without SToP, using a fixed 5-token emission per frame; (2) without FDTC, reverting to LiveStar’s response trigger while keeping SToP; and (3) LiveStar backbone without control modules.
As shown in Tab.~\ref{tab:ablation_control}, removing FDTC causes the largest drop: the model collapses to two output states (T/S), losing the \textbf{Continuing State} that is the actual novel ingredient over the backbone's single-pass verification. Without it, LyraV emits short and incomplete sentences, which severely degrades semantic score (SS -34.4\%), narrative flow (NF -55.4\%), and increases response latency (RL +15.4\%) due to poor emit/silence decisions. This confirms that the quality gains attributed to FDTC stem specifically from prefix-level continuation, not from the verification mechanism inherited from LiveStar.
Removing SToP, by contrast, leaves caption quality almost intact (SS -8.6\%, NF -7.9\%): a fixed-rate emission can already approach full quality because the latency cutoff bounds emission regardless. We read this as evidence \emph{consistent with} our positioning rather than against it---SToP is not meant to raise caption-level scores but to regulate \emph{how many} tokens are emitted per frame so that playback stays stutter-free; its load-bearing effect is therefore on streaming fluidity and throughput (Tab.~\ref{tab:pacer_comparison}, Tab.~\ref{tab:synchrony_exp}), not on SS/NF.

\noindent\textbf{Analysis of Pacer Architecture.}
The SToP module predicts per-frame token generation rates using a short history ($N$ = 4 frames). To assess architectural impact, we compare its Transformer encoder with RNN~\cite{rumelhart1985learning} and LSTM~\cite{hochreiter1997long} variants under identical settings. As shown in Tab.~\ref{tab:pacer_comparison}, Transformer achieves the highest prediction accuracy and system fluidity (rFPS). Despite greater theoretical complexity, its lightweight two-layer design and parallel computation enable faster and more accurate performance than sequential RNN-based models. These results validate that Transformer Pacer delivers the optimal balance of precision and efficiency for short-sequence, low-latency pacing prediction. We provide a detailed analysis of Pacer's implementation, efficiency, and frame rate generalization in Sec~\ref{analysis_pacer}.

\begin{figure}[t]
\centering
\includegraphics[width=\linewidth]{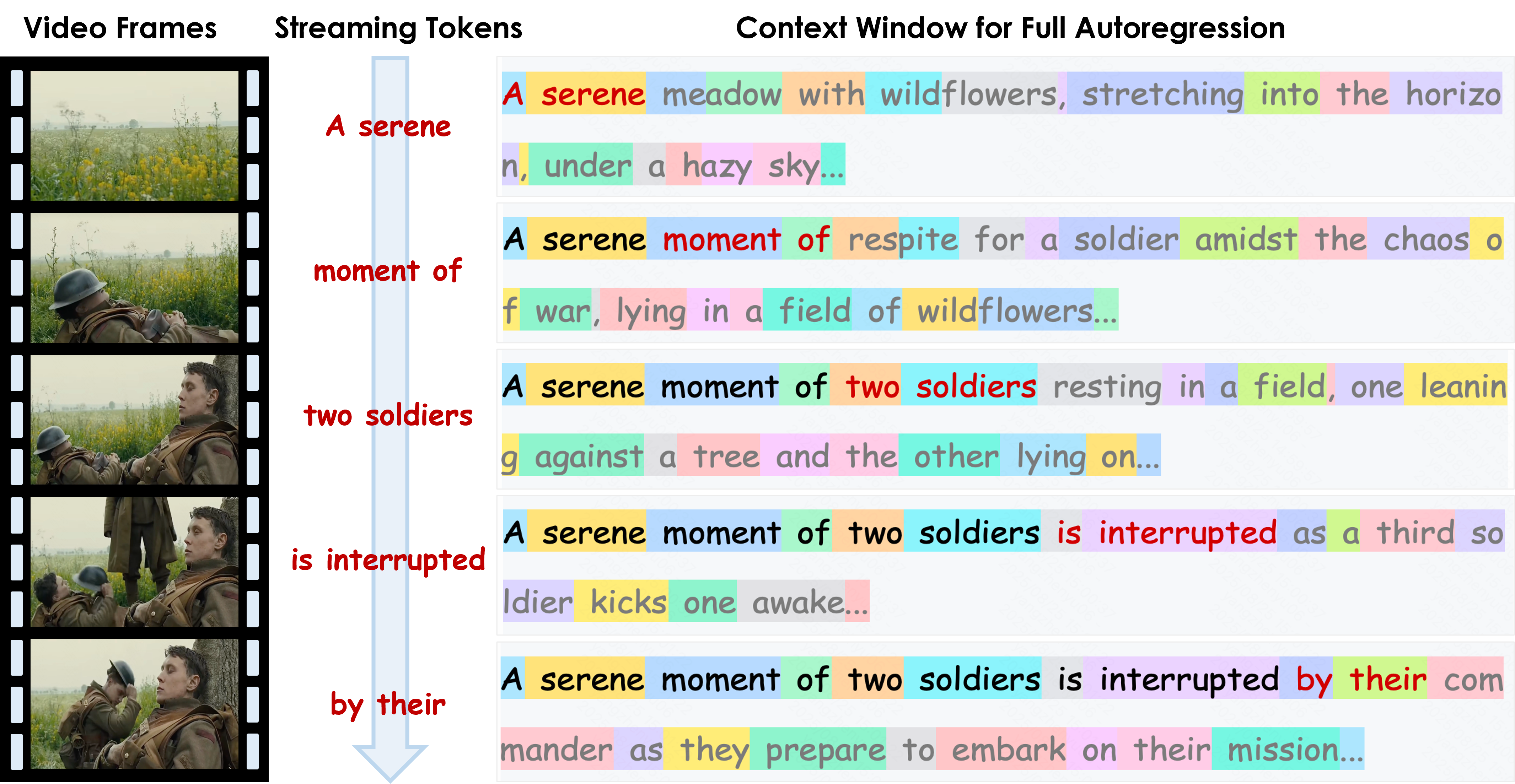}
\caption{\small \textbf{Case Study.} Fixed prediction (2 tokens/frame) with full context visualization. \textcolor{gray}{Gray tokens} are non-decoded "thoughts".}
\label{fig:streaming_case}
\end{figure}
\subsection{Observation: Dynamic Reasoning over Streaming Tokens}
\label{subsec:insight}
During inference, we qualitatively observe that LyraV tends to interpret and refine its narration alongside the visual input, rather than committing to a fixed description from a single trigger frame. We stress that this is a training-free behavioral observation drawn from our demo and case studies, not a quantitatively validated capability. In a case study (Fig. \ref{fig:streaming_case}), we fix SToP prediction \( \hat{m}_i = 2 \) (decoding only two tokens per frame) while visualizing the full autoregressive context. Gray tokens denote internal non-decoded continuations that the model would produce at each step, which we use to visualize how its tentative interpretation evolves over time.
As frames progress, LyraV refines its interpretation incrementally—e.g., transitioning from "A serene meadow..." to "A serene moment... a soldier...", and finally "A serene moment of two soldiers...". This qualitative example suggests that LyraV processes content incrementally and may offer potential advantages in low-latency inference and error tolerance compared to Video-LLMs that require full decoding; we leave a quantitative study of this behavior to future work.

\begin{figure*}[t]
\centering
\includegraphics[width=0.95\textwidth]{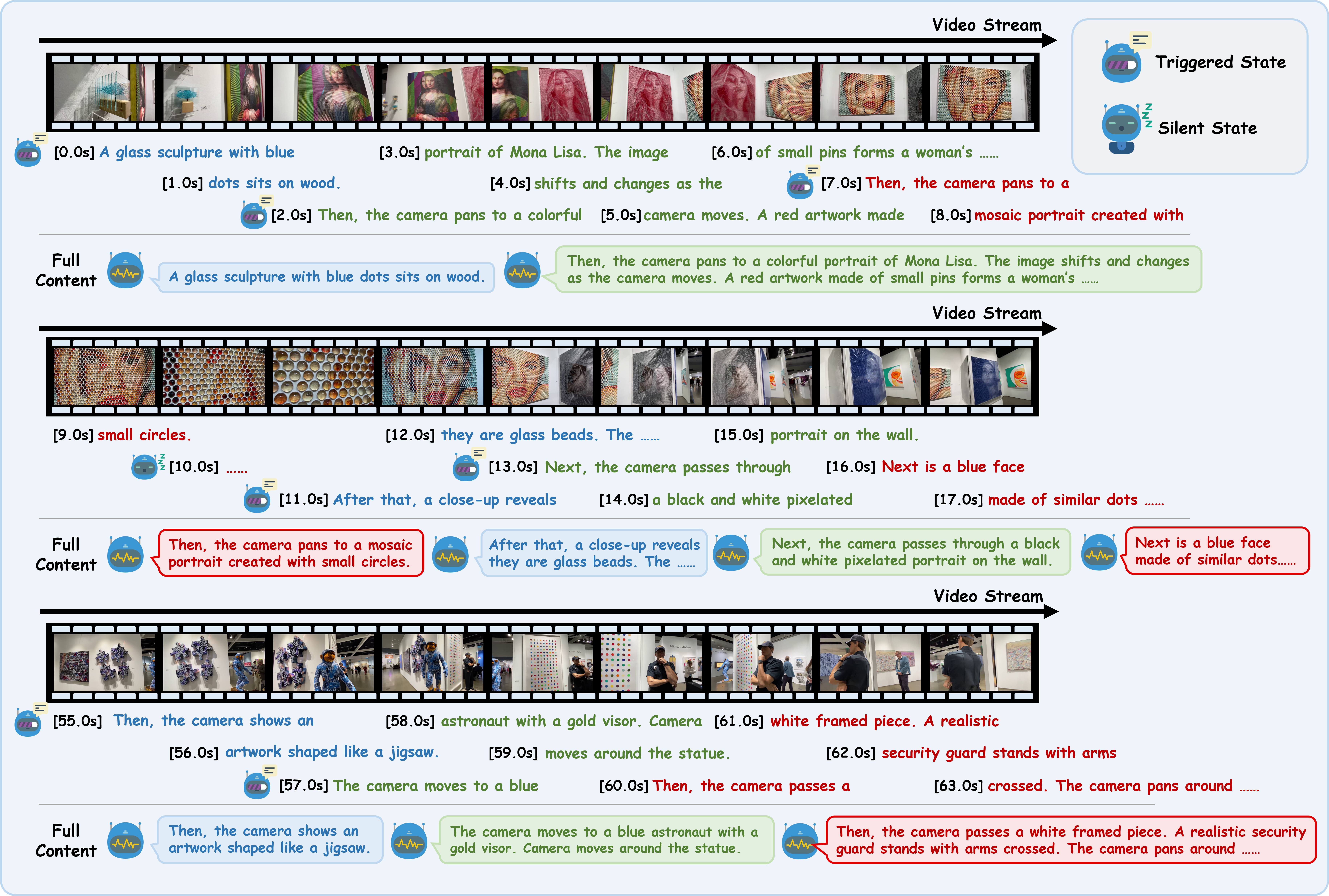}
\caption{\small \textbf{A qualitative visualization of LyraV's real-time narration process.} LyraV can generate descriptive content synchronously with video playback, vividly
illustrating the core mechanism of Streaming Video-Language
Synchrony (SVLS).} 
\label{fig:case_study_lyrav_1}
\end{figure*}

\section{Discussion and Analysis}
\subsection{Performance Analysis}
Due to paradigm differences, our main advantage lies in synchronization and fluency rather than static accuracy.
On QA benchmarks (passive tasks with short answers), our results indicate that LyraV preserves the backbone's understanding capabilities without notable degradation. In Real-time Captioning (proactive tasks), LyraV shows consistent improvements in semantic scores and latency. These gains are most pronounced on our Synchrony Benchmarks, where LyraV attains higher synchrony while maintaining comparable semantic quality.
We note that the Sync Rate itself is, by construction, sensitive to a model's truncation policy and should not be interpreted as a standalone quality signal; accordingly, the load-bearing evidence for our claims is the content quality (SS/NF) measured under the truncated, equal-budget protocol, where all models share an identical 100\% SR and the only remaining difference is narration quality.
These results are consistent with the dynamic reasoning behavior we observe qualitatively: existing proactive models generate responses from a single static trigger frame and thus may miss subsequent micro-events, whereas LyraV continues to attend to the evolving visual stream within a single caption.
Regarding baselines, our streaming vision–language synchrony task requires proactive response capabilities. Thus, methods such as Flash-VStream and ReKV, which lack such mechanisms, are not applicable in our active setting. We instead evaluate the proactive model LiveCC under the same synchrony setup, where LyraV improves over it in SR, SS, and NF in Tab.~\ref{tab:online_exp}. We further separately evaluate real-time processing speed (rFPS) under the online evaluation, where LyraV is also slightly faster than LiveCC (3.89 vs.\ 3.77).

\subsection{Theoretical Basis of FDTC Decision Rule}
The decision rule via the PPL-based verification mechanism in FDTC is well-designed and thoroughly validated in LiveStar. Intuitively, PPL measures the model's uncertainty in generating a specific text sequence given a context. By computing the PPL of the \textit{ongoing caption} conditioned on the \textit{new visual input}, we assess whether the current narration is still valid. A stable PPL indicates that the caption remains aligned with the evolving visual stream (i.e., the model would likely generate this same content even if decoding from scratch). Conversely, a PPL surge signals a semantic mismatch or event boundary where the previous context no longer fits.
In FDTC, when a new frame arrives during an ongoing response (in truncated synchrony setting), we halt the current decoding and verify the accumulated context using this criterion. If the PPL remains stable relative to the previous step, the model enters the Continuing State, where the current context with updated visual features is directly input to generate subsequent tokens. If the PPL surges, the model enters the Triggered State: we append the termination token \texttt{<|im\_end|>} to conclude the current sentence and force a new prefix \texttt{<|im\_start|>assistant} to initiate a fresh caption.

\begin{figure*}[t]
\centering
\includegraphics[width=0.95\textwidth]{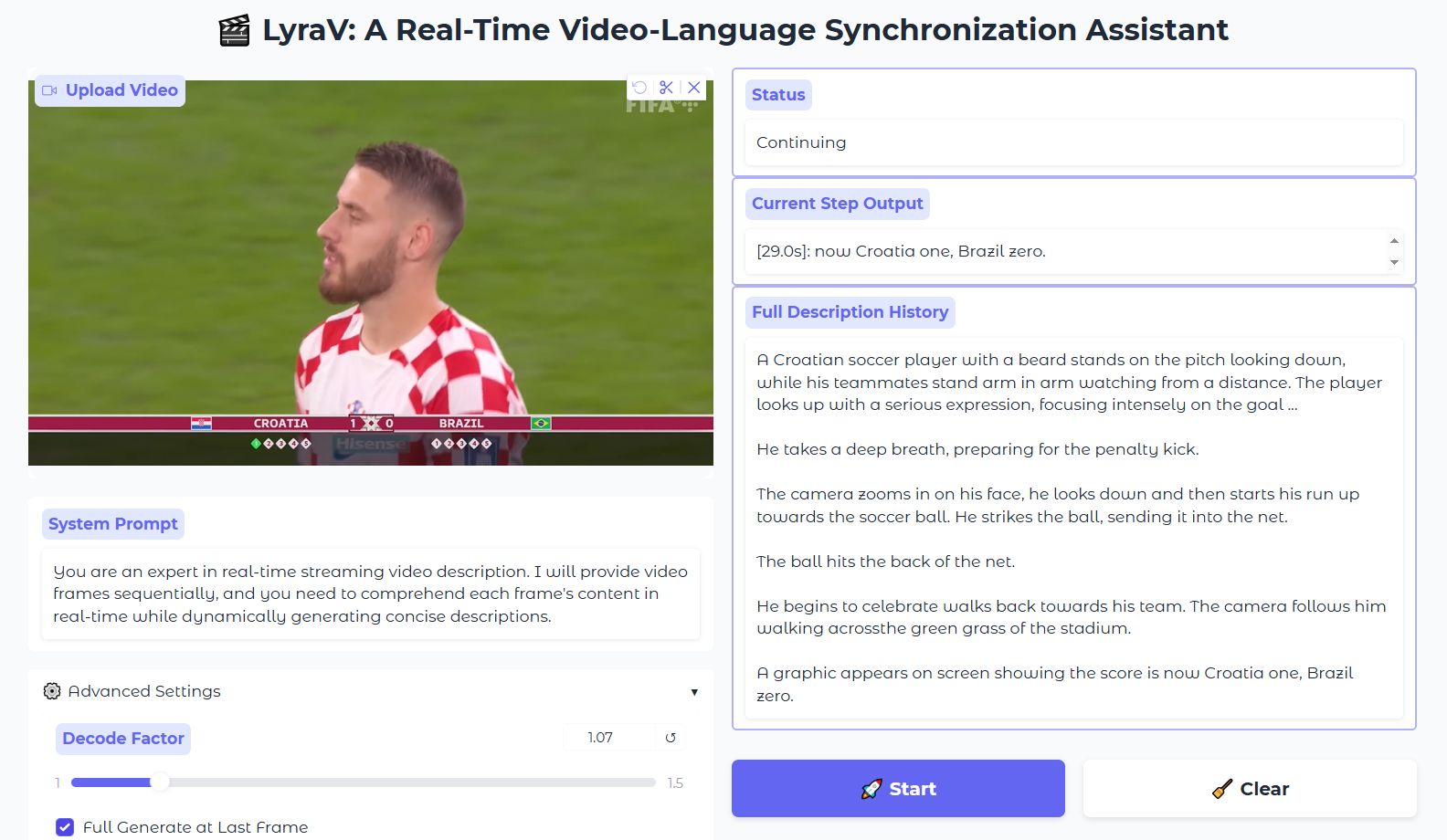}
\caption{\small \textbf{An illustration of the LyraV web demo interface.} The model processes a soccer match video in real-time, displaying its current state, the incremental output for the latest frame, and the accumulated full description. Users can also adjust inference parameters like the Decode Factor \( \alpha \) in the advanced settings.}
\label{fig:LyraV_visualization}
\end{figure*}

\subsection{Technical Novelty of FDTC and SToP}
Our core innovation lies in the algorithmic logic and control flow of the \textbf{Frame-Driven Transition Controller (FDTC)}. Unlike LiveStar's verification mechanism, which performs a binary ``speak or wait'' check on a full caption in a single pass, FDTC implements an incremental, stateful mechanism that verifies partial, ongoing sentence prefixes relative to incoming frames. This capability enables the novel \textbf{Continuing State} achieving smooth synchronization. 
Regarding the \textbf{Streaming Token Pacer (SToP)}, to our knowledge, no prior work employs a module to predict speech rate based on real-time visual information density and to reconcile it with a hard latency budget. We position SToP's contribution as a \emph{problem formulation} (content-aware, latency-bounded pacing of token emission) together with a practical scheduling component, rather than as a source of caption-level accuracy. Consistent with our ablation, its value is realized as system-level fluidity under tight budgets; its supervision (human speaking rate) is admittedly a weak, noisy proxy (Sec.~\ref{sec:FDTC}), which is precisely why it is used as a soft prior subordinate to the latency cutoff. FDTC and SToP remain complementary: state transitions decide \emph{when} and \emph{whether} to speak, while SToP regulates \emph{how fast} under real-time constraints.

\subsection{Threshold Sensitivity Analysis}
When $\alpha$ is too low, the model becomes hypersensitive, causing unstable state transitions and frequent triggering ($T \leftrightarrow C$), which leads to fragmented and incomplete captions. When $\alpha$ is too high, new responses are rarely triggered (stay $S$): the model overextends the current caption until EOS and often remains stuck in the Silent State. Both extremes correspond to typical failure cases of streaming narration. We find $\alpha \approx 1.05$ achieves the best balance.

\subsection{Analysis of Pacer's Implementation, Efficiency, and Frame Rate Generalization}
\label{analysis_pacer}
We refer to the SToP module as ``lightweight'' due to its minimal computational overhead relative to the Video-LLM. As shown in Tab.~\ref{tab:synchrony_exp}, SToP runs at $\sim$200 FPS and adds negligible inference latency. In practice, the pacer is implemented as a compact Transformer operating in a reduced latent space. Visual features extracted from the video encoder ($D_v=4096$) are first aggregated into frame-level representations via mean pooling across visual tokens and then projected to a lower-dimensional space with $d=512$. The token-count embeddings are projected to the same latent space, and the resulting sequence is processed by a Transformer encoder with two layers and hidden dimension $d=512$, followed by a two-layer MLP for token-budget prediction. Regarding frame rate generalization, although trained at 2fps, SToP generalizes well to other frame rates (e.g., 1fps or 4fps) by proportionally scaling the predicted token count (e.g., $\times 2$ or $\times 0.5$) to match the time budget, ensuring robust pacing.

\section{Visualization of LyraV}
\subsection{Case Study}
Figure \ref{fig:case_study_lyrav_1} provides a qualitative visualization of LyraV's dynamic narration process on a continuous video, vividly illustrating the core mechanism of Streaming Video-Language Synchrony. As depicted, rather than waiting for an event to conclude before generating a full sentence, LyraV seamlessly interleaves generated token fragments with the incoming video frames, achieving a "speak-while-watching" capability. Our hierarchical control framework plays a crucial role in this process. The FDTC module, acting as the high-level controller, intelligently punctuates the narrative. It enters a Triggered State to initiate a new utterance only when a significant change in visual content is detected (e.g., at 7.0s when the camera pans to a new artwork, or at 11.0s for a close-up). During intervals of relative semantic stability (e.g., at 10.0s or 56.0s), it enters a Silent State, thereby effectively avoiding redundant descriptions. Concurrently, the SToP module handles the fine-grained generation, ensuring the number of tokens produced at each time step matches the visual rhythm. Ultimately, these dynamically generated and seemingly discrete fragments are composed in real-time into a fluid, coherent, and highly synchronized full-content narrative.

\subsection{Web Demo}

We have built an interactive web demo for LyraV using Gradio, allowing users to upload any video and observe the real-time generation of synchronized narrative descriptions. As the model processes each frame, it not only generates corresponding tokens but also continuously updates the full utterance buffer. The interface provides a real-time view of the model's internal state (e.g., “Continuing”), the incremental output for the current time step, and the complete description history. Furthermore, it supports dynamic configuration of key inference parameters (such as setting a fixed number of decoding tokens or enabling real-time latency adaptation, along with adjustable scaling factors like the Decode Factor \( \alpha \), which controls the sensitivity of the FDTC's triggering mechanism to balance responsiveness and narrative completeness) to suit diverse use cases. A preview of the demo interface is shown in Figure~\ref{fig:LyraV_visualization}.

\section{Conclusion}
In this paper, we introduce Streaming Video-Language Synchrony (SVLS), a paradigm that lets Video-LLMs interleave perception and generation at frame–token granularity instead of pausing perception to decode full sentences. We realize SVLS through LyraV, a lightweight control layer wrapped around a frozen online Video-LLM, comprising the Frame-Driven Transition Controller (FDTC) and the Streaming Token Pacer (SToP). Our experiments indicate that LyraV preserves the frozen backbone's general video understanding while improving streaming synchrony and narration quality under real-time constraints; we emphasize that its contribution lies primarily in the synchrony paradigm and control framework rather than in raising static, single-answer accuracy. Through an interactive demo and case studies, we further note a qualitative incremental-refinement behavior during streaming, where LyraV updates its narration as visual evidence accumulates; we present this as an empirical observation and leave its rigorous quantification to future work.

\bibliographystyle{IEEEtran}
\bibliography{references}

\begin{thebibliography}{100}
\providecommand{\url}[1]{#1}
\csname url@samestyle\endcsname
\providecommand{\newblock}{\relax}
\providecommand{\bibinfo}[2]{#2}
\providecommand{\BIBentrySTDinterwordspacing}{\spaceskip=0pt\relax}
\providecommand{\BIBentryALTinterwordstretchfactor}{4}
\providecommand{\BIBentryALTinterwordspacing}{\spaceskip=\fontdimen2\font plus
\BIBentryALTinterwordstretchfactor\fontdimen3\font minus \fontdimen4\font\relax}
\providecommand{\BIBforeignlanguage}[2]{{%
\expandafter\ifx\csname l@#1\endcsname\relax
\typeout{** WARNING: IEEEtran.bst: No hyphenation pattern has been}%
\typeout{** loaded for the language `#1'. Using the pattern for}%
\typeout{** the default language instead.}%
\else
\language=\csname l@#1\endcsname
\fi
#2}}
\providecommand{\BIBdecl}{\relax}
\BIBdecl

\bibitem{ataallah2024minigpt4}
K.~Ataallah, X.~Shen, E.~Abdelrahman, E.~Sleiman, D.~Zhu, J.~Ding, and M.~Elhoseiny, ``Minigpt4-video: Advancing multimodal llms for video understanding with interleaved visual-textual tokens,'' \emph{arXiv preprint arXiv:2404.03413}, 2024.

\bibitem{maaz2023video}
M.~Maaz, H.~Rasheed, S.~Khan, and F.~S. Khan, ``Video-chatgpt: Towards detailed video understanding via large vision and language models,'' \emph{arXiv preprint arXiv:2306.05424}, 2023.

\bibitem{li2023videochat}
K.~Li, Y.~He, Y.~Wang, Y.~Li, W.~Wang, P.~Luo, Y.~Wang, L.~Wang, and Y.~Qiao, ``Videochat: Chat-centric video understanding,'' \emph{arXiv preprint arXiv:2305.06355}, 2023.

\bibitem{yang2023vid2seq}
A.~Yang, A.~Nagrani, P.~H. Seo, A.~Miech, J.~Pont-Tuset, I.~Laptev, J.~Sivic, and C.~Schmid, ``Vid2seq: Large-scale pretraining of a visual language model for dense video captioning,'' in \emph{Proceedings of the IEEE/CVF Conference on Computer Vision and Pattern Recognition}, 2023, pp. 10\,714--10\,726.

\bibitem{wang2022internvideo}
Y.~Wang, K.~Li, Y.~Li, Y.~He, B.~Huang, Z.~Zhao, H.~Zhang, J.~Xu, Y.~Liu, Z.~Wang \emph{et~al.}, ``Internvideo: General video foundation models via generative and discriminative learning,'' \emph{arXiv preprint arXiv:2212.03191}, 2022.

\bibitem{11050020}
Q.~Ye, Z.~Yu, R.~Shao, Y.~Cui, X.~Kang, X.~Liu, P.~Torr, and X.~Cao, ``Cat+: Investigating and enhancing audio-visual understanding in large language models,'' \emph{IEEE Transactions on Pattern Analysis and Machine Intelligence}, vol.~47, no.~10, pp. 8674--8690, 2025.

\bibitem{11223149}
L.-H. Chen, S.~Lu, A.~Zeng, H.~Zhang, B.~Wang, R.~Zhang, and L.~Zhang, ``Motionllm: Understanding human behaviors from human motions and videos,'' \emph{IEEE Transactions on Pattern Analysis and Machine Intelligence}, pp. 1--15, 2025.

\bibitem{cheng2024videollama}
Z.~Cheng, S.~Leng, H.~Zhang, Y.~Xin, X.~Li, G.~Chen, Y.~Zhu, W.~Zhang, Z.~Luo, D.~Zhao \emph{et~al.}, ``Videollama 2: Advancing spatial-temporal modeling and audio understanding in video-llms,'' \emph{arXiv preprint arXiv:2406.07476}, 2024.

\bibitem{liu2024oryx}
Z.~Liu, Y.~Dong, Z.~Liu, W.~Hu, J.~Lu, and Y.~Rao, ``Oryx mllm: On-demand spatial-temporal understanding at arbitrary resolution,'' \emph{arXiv preprint arXiv:2409.12961}, 2024.

\bibitem{ren2024timechat}
S.~Ren, L.~Yao, S.~Li, X.~Sun, and L.~Hou, ``Timechat: A time-sensitive multimodal large language model for long video understanding,'' in \emph{Proceedings of the IEEE/CVF Conference on Computer Vision and Pattern Recognition}, 2024, pp. 14\,313--14\,323.

\bibitem{11202655}
S.~A. Peirone, F.~Pistilli, A.~Alliegro, T.~Tommasi, and G.~Averta, ``Hier-egopack: Hierarchical egocentric video understanding with diverse task perspectives,'' \emph{IEEE Transactions on Pattern Analysis and Machine Intelligence}, vol.~48, no.~2, pp. 1917--1931, 2026.

\bibitem{zhang2024flash}
H.~Zhang, Y.~Wang, Y.~Tang, Y.~Liu, J.~Feng, J.~Dai, and X.~Jin, ``Flash-vstream: Memory-based real-time understanding for long video streams,'' \emph{arXiv preprint arXiv:2406.08085}, 2024.

\bibitem{song2024moviechat+}
E.~Song, W.~Chai, T.~Ye, J.-N. Hwang, X.~Li, and G.~Wang, ``Moviechat+: Question-aware sparse memory for long video question answering,'' \emph{arXiv preprint arXiv:2404.17176}, 2024.

\bibitem{he2024ma}
B.~He, H.~Li, Y.~K. Jang, M.~Jia, X.~Cao, A.~Shah, A.~Shrivastava, and S.-N. Lim, ``Ma-lmm: Memory-augmented large multimodal model for long-term video understanding,'' in \emph{Proceedings of the IEEE/CVF Conference on Computer Vision and Pattern Recognition}, 2024, pp. 13\,504--13\,514.

\bibitem{wang2024longllavascalingmultimodalllms}
\BIBentryALTinterwordspacing
X.~Wang, D.~Song, S.~Chen, C.~Zhang, and B.~Wang, ``Longllava: Scaling multi-modal llms to 1000 images efficiently via hybrid architecture,'' 2024. [Online]. Available: \url{https://arxiv.org/abs/2409.02889}
\BIBentrySTDinterwordspacing

\bibitem{longvila}
F.~Xue, Y.~Chen, D.~Li, Q.~Hu, L.~Zhu, X.~Li, Y.~Fang, H.~Tang, S.~Yang, Z.~Liu, Y.~He, H.~Yin, P.~Molchanov, J.~Kautz, L.~Fan, Y.~Zhu, Y.~Lu, and S.~Han, ``Longvila: Scaling long-context visual language models for long videos,'' \emph{null}, 2024.

\bibitem{zhang2024long}
P.~Zhang, K.~Zhang, B.~Li, G.~Zeng, J.~Yang, Y.~Zhang, Z.~Wang, H.~Tan, C.~Li, and Z.~Liu, ``Long context transfer from language to vision,'' \emph{arXiv preprint arXiv:2406.16852}, 2024.

\bibitem{11359544}
J.~Li, M.~Gao, X.~He, S.~Tang, W.-S. Zheng, J.~Xiao, M.~Wang, T.-S. Chua, and Y.~Zhuang, ``Momentor++: Advancing video large language models with fine-grained long video reasoning,'' \emph{IEEE Transactions on Pattern Analysis and Machine Intelligence}, vol.~48, no.~6, pp. 6208--6224, 2026.

\bibitem{11430664}
K.~Zhang, Z.~Yang, M.~Han, Y.~Zhuge, H.~Hao, C.~Li, Z.~Li, and X.~Chang, ``Selongvlm: Empowering long video language models with self-corrective clip selection,'' \emph{IEEE Transactions on Pattern Analysis and Machine Intelligence}, pp. 1--16, 2026.

\bibitem{11535730}
S.~Tian, R.~Wang, H.~Guo, P.~Wu, Y.~Dong, X.~Wang, J.~Yang, H.~Zhang, H.~Zhu, and Z.~Liu, ``Ego-r1: Agentic chain-of-tool-thought for ultra-long egocentric video reasoning,'' \emph{IEEE Transactions on Pattern Analysis and Machine Intelligence}, pp. 1--16, 2026.

\bibitem{lee2018interaction}
L.-H. Lee and P.~Hui, ``Interaction methods for smart glasses: A survey,'' \emph{IEEE access}, vol.~6, pp. 28\,712--28\,732, 2018.

\bibitem{sutherland1968head}
I.~E. Sutherland, ``A head-mounted three dimensional display,'' in \emph{Proceedings of the December 9-11, 1968, fall joint computer conference, part I}, 1968, pp. 757--764.

\bibitem{horn1986robot}
B.~Horn, \emph{Robot vision}.\hskip 1em plus 0.5em minus 0.4em\relax MIT press, 1986.

\bibitem{chen2024videollm}
J.~Chen, Z.~Lv, S.~Wu, K.~Q. Lin, C.~Song, D.~Gao, J.-W. Liu, Z.~Gao, D.~Mao, and M.~Z. Shou, ``Videollm-online: Online video large language model for streaming video,'' in \emph{Proceedings of the IEEE/CVF Conference on Computer Vision and Pattern Recognition}, 2024, pp. 18\,407--18\,418.

\bibitem{wu2024videollm}
S.~Wu, J.~Chen, K.~Q. Lin, Q.~Wang, Y.~Gao, Q.~Xu, T.~Xu, Y.~Hu, E.~Chen, and M.~Z. Shou, ``Videollm-mod: Efficient video-language streaming with mixture-of-depths vision computation,'' \emph{Advances in Neural Information Processing Systems}, vol.~37, pp. 109\,922--109\,947, 2024.

\bibitem{li2025lion}
W.~Li, B.~Hu, R.~Shao, L.~Shen, and L.~Nie, ``Lion-fs: Fast \& slow video-language thinker as online video assistant,'' \emph{arXiv preprint arXiv:2503.03663}, 2025.

\bibitem{ding2025streammind}
X.~Ding, H.~Wu, Y.~Yang, S.~Jiang, D.~Bai, Z.~Chen, and T.~Cao, ``Streammind: Unlocking full frame rate streaming video dialogue through event-gated cognition,'' \emph{arXiv preprint arXiv:2503.06220}, 2025.

\bibitem{wang2024videollm}
Y.~Wang, X.~Meng, Y.~Wang, J.~Liang, J.~Wei, H.~Zhang, and D.~Zhao, ``Videollm knows when to speak: Enhancing time-sensitive video comprehension with video-text duet interaction format,'' \emph{arXiv preprint arXiv:2411.17991}, 2024.

\bibitem{wang2025streambridge}
H.~Wang, B.~Feng, Z.~Lai, M.~Xu, S.~Li, W.~Ge, A.~Dehghan, M.~Cao, and P.~Huang, ``Streambridge: Turning your offline video large language model into a proactive streaming assistant,'' \emph{arXiv preprint arXiv:2505.05467}, 2025.

\bibitem{qian2024streaming}
R.~Qian, X.~Dong, P.~Zhang, Y.~Zang, S.~Ding, D.~Lin, and J.~Wang, ``Streaming long video understanding with large language models,'' \emph{Advances in Neural Information Processing Systems}, vol.~37, pp. 119\,336--119\,360, 2024.

\bibitem{yang2025livestar}
\BIBentryALTinterwordspacing
Z.~Yang, K.~Zhang, Y.~Hu, B.~Wang, S.~Qian, B.~Wen, F.~Yang, T.~Gao, W.~Dong, and C.~Xu, ``Livestar: Live streaming assistant for real-world online video understanding,'' in \emph{The Thirty-ninth Annual Conference on Neural Information Processing Systems}, 2025. [Online]. Available: \url{https://openreview.net/forum?id=4n7IifN7yr}
\BIBentrySTDinterwordspacing

\bibitem{bai2025speakstream}
R.~H. Bai, Z.~Gu, T.~Likhomanenko, and N.~Jaitly, ``Speakstream: Streaming text-to-speech with interleaved data,'' \emph{arXiv preprint arXiv:2505.19206}, 2025.

\bibitem{giancola2018soccernet}
S.~Giancola, M.~Amine, T.~Dghaily, and B.~Ghanem, ``Soccernet: A scalable dataset for action spotting in soccer videos,'' in \emph{Proceedings of the IEEE conference on computer vision and pattern recognition workshops}, 2018, pp. 1711--1721.

\bibitem{taniguchi2019generating}
Y.~Taniguchi, Y.~Feng, H.~Takamura, and M.~Okumura, ``Generating live soccer-match commentary from play data,'' in \emph{Proceedings of the AAAI Conference on Artificial Intelligence}, vol.~33, no.~01, 2019, pp. 7096--7103.

\bibitem{chen2024sharegpt4video}
L.~Chen, X.~Wei, J.~Li, X.~Dong, P.~Zhang, Y.~Zang, Z.~Chen, H.~Duan, Z.~Tang, L.~Yuan \emph{et~al.}, ``Sharegpt4video: Improving video understanding and generation with better captions,'' \emph{Advances in Neural Information Processing Systems}, vol.~37, pp. 19\,472--19\,495, 2024.

\bibitem{xu2024pllava}
L.~Xu, Y.~Zhao, D.~Zhou, Z.~Lin, S.~K. Ng, and J.~Feng, ``Pllava: Parameter-free llava extension from images to videos for video dense captioning,'' \emph{arXiv preprint arXiv:2404.16994}, 2024.

\bibitem{islam2024video}
M.~M. Islam, N.~Ho, X.~Yang, T.~Nagarajan, L.~Torresani, and G.~Bertasius, ``Video recap: Recursive captioning of hour-long videos,'' in \emph{Proceedings of the IEEE/CVF Conference on Computer Vision and Pattern Recognition}, 2024, pp. 18\,198--18\,208.

\bibitem{touvron2023llama}
H.~Touvron, L.~Martin, K.~Stone, P.~Albert, A.~Almahairi, Y.~Babaei, N.~Bashlykov, S.~Batra, P.~Bhargava, S.~Bhosale \emph{et~al.}, ``Llama 2: Open foundation and fine-tuned chat models,'' \emph{arXiv preprint arXiv:2307.09288}, 2023.

\bibitem{team2023gemini}
G.~Team, R.~Anil, S.~Borgeaud, J.-B. Alayrac, J.~Yu, R.~Soricut, J.~Schalkwyk, A.~M. Dai, A.~Hauth, K.~Millican \emph{et~al.}, ``Gemini: a family of highly capable multimodal models,'' \emph{arXiv preprint arXiv:2312.11805}, 2023.

\bibitem{achiam2023gpt}
J.~Achiam, S.~Adler, S.~Agarwal, L.~Ahmad, I.~Akkaya, F.~L. Aleman, D.~Almeida, J.~Altenschmidt, S.~Altman, S.~Anadkat \emph{et~al.}, ``Gpt-4 technical report,'' \emph{arXiv preprint arXiv:2303.08774}, 2023.

\bibitem{ouyang2022training}
L.~Ouyang, J.~Wu, X.~Jiang, D.~Almeida, C.~Wainwright, P.~Mishkin, C.~Zhang, S.~Agarwal, K.~Slama, A.~Ray \emph{et~al.}, ``Training language models to follow instructions with human feedback,'' \emph{Advances in neural information processing systems}, vol.~35, pp. 27\,730--27\,744, 2022.

\bibitem{radford2018improving}
A.~Radford, K.~Narasimhan, T.~Salimans, I.~Sutskever \emph{et~al.}, ``Improving language understanding by generative pre-training,'' \emph{Unknown}, 2018.

\bibitem{yang2024ldre}
Z.~Yang, D.~Xue, S.~Qian, W.~Dong, and C.~Xu, ``Ldre: Llm-based divergent reasoning and ensemble for zero-shot composed image retrieval,'' in \emph{Proceedings of the 47th International ACM SIGIR conference on research and development in information retrieval}, 2024, pp. 80--90.

\bibitem{lin2024vila}
J.~Lin, H.~Yin, W.~Ping, P.~Molchanov, M.~Shoeybi, and S.~Han, ``Vila: On pre-training for visual language models,'' in \emph{Proceedings of the IEEE/CVF Conference on Computer Vision and Pattern Recognition}, 2024, pp. 26\,689--26\,699.

\bibitem{yang2024semantic}
Z.~Yang, S.~Qian, D.~Xue, J.~Wu, F.~Yang, W.~Dong, and C.~Xu, ``Semantic editing increment benefits zero-shot composed image retrieval,'' in \emph{Proceedings of the 32nd ACM International Conference on Multimedia}, 2024, pp. 1245--1254.

\bibitem{ko2023large}
D.~Ko, J.~S. Lee, W.~Kang, B.~Roh, and H.~J. Kim, ``Large language models are temporal and causal reasoners for video question answering,'' \emph{arXiv preprint arXiv:2310.15747}, 2023.

\bibitem{li2024mvbench}
K.~Li, Y.~Wang, Y.~He, Y.~Li, Y.~Wang, Y.~Liu, Z.~Wang, J.~Xu, G.~Chen, P.~Luo \emph{et~al.}, ``Mvbench: A comprehensive multi-modal video understanding benchmark,'' in \emph{Proceedings of the IEEE/CVF Conference on Computer Vision and Pattern Recognition}, 2024, pp. 22\,195--22\,206.

\bibitem{maaz2024videogpt+}
M.~Maaz, H.~Rasheed, S.~Khan, and F.~Khan, ``Videogpt+: Integrating image and video encoders for enhanced video understanding,'' \emph{arXiv preprint arXiv:2406.09418}, 2024.

\bibitem{zhang2024llavanext-video}
\BIBentryALTinterwordspacing
Y.~Zhang, B.~Li, h.~Liu, Y.~j. Lee, L.~Gui, D.~Fu, J.~Feng, Z.~Liu, and C.~Li, ``Llava-next: A strong zero-shot video understanding model,'' April 2024. [Online]. Available: \url{https://llava-vl.github.io/blog/2024-04-30-llava-next-video/}
\BIBentrySTDinterwordspacing

\bibitem{9770842}
A.~Yang, A.~Miech, J.~Sivic, I.~Laptev, and C.~Schmid, ``Learning to answer visual questions from web videos,'' \emph{IEEE Transactions on Pattern Analysis and Machine Intelligence}, vol.~47, no.~5, pp. 3202--3218, 2025.

\bibitem{10214041}
Y.~Li, X.~Wang, J.~Xiao, W.~Ji, and T.-S. Chua, ``Transformer-empowered invariant grounding for video question answering,'' \emph{IEEE Transactions on Pattern Analysis and Machine Intelligence}, vol.~47, no.~11, pp. 9510--9522, 2025.

\bibitem{11506215}
J.~Li, P.~Wei, W.~Han, S.-C. Zhu, and L.~Fan, ``Intentqa: Intent question answering in videos by cognitive context reasoning,'' \emph{IEEE Transactions on Pattern Analysis and Machine Intelligence}, pp. 1--18, 2026.

\bibitem{11329152}
J.~Li, Z.~Liao, F.~Xiao, T.~Li, Q.~Zhang, H.~Zhao, L.~Niu, G.~Chen, L.~Zhang, and C.~Jiang, ``Parse, align and aggregate: Graph-driven compositional reasoning for video question answering,'' \emph{IEEE Transactions on Pattern Analysis and Machine Intelligence}, vol.~48, no.~5, pp. 5586--5603, 2026.

\bibitem{11219357}
T.~Chen, H.~Liu, Y.~Wang, Y.~Chen, T.~He, C.~Gan, H.~He, and W.~Lin, ``Mecd+: Unlocking event-level causal graph discovery for video reasoning,'' \emph{IEEE Transactions on Pattern Analysis and Machine Intelligence}, vol.~48, no.~3, pp. 2628--2645, 2026.

\bibitem{11391656}
L.-L. Li, J.~Fang, J.~Xiao, H.~Yu, C.~Lv, J.~Xue, Z.~Li, and T.-S. Chua, ``Adversa: Abductive driving accident video understanding,'' \emph{IEEE Transactions on Pattern Analysis and Machine Intelligence}, vol.~48, no.~6, pp. 6980--6998, 2026.

\bibitem{11146594}
E.~Song, W.~Chai, T.~Ye, J.-N. Hwang, X.~Li, and G.~Wang, ``Moviechat+: Question-aware sparse memory for long video question answering,'' \emph{IEEE Transactions on Pattern Analysis and Machine Intelligence}, vol.~48, no.~1, pp. 374--389, 2026.

\bibitem{guo2025vtg}
Y.~Guo, J.~Liu, M.~Li, D.~Cheng, X.~Tang, D.~Sui, Q.~Liu, X.~Chen, and K.~Zhao, ``Vtg-llm: Integrating timestamp knowledge into video llms for enhanced video temporal grounding,'' in \emph{Proceedings of the AAAI Conference on Artificial Intelligence}, vol.~39, no.~3, 2025, pp. 3302--3310.

\bibitem{xu2024vtg}
Y.~Xu, Y.~Sun, Z.~Xie, B.~Zhai, and S.~Du, ``Vtg-gpt: Tuning-free zero-shot video temporal grounding with gpt,'' \emph{Applied Sciences}, vol.~14, no.~5, p. 1894, 2024.

\bibitem{wang2024hawkeye}
Y.~Wang, X.~Meng, J.~Liang, Y.~Wang, Q.~Liu, and D.~Zhao, ``Hawkeye: Training video-text llms for grounding text in videos,'' \emph{arXiv preprint arXiv:2403.10228}, 2024.

\bibitem{11184436}
J.~Wu, W.~Liu, Y.~Liu, M.~Liu, L.~Nie, Z.~Lin, and C.~W. Chen, ``A survey on video temporal grounding with multimodal large language model,'' \emph{IEEE Transactions on Pattern Analysis and Machine Intelligence}, vol.~48, no.~2, pp. 1521--1541, 2026.

\bibitem{yang2023dawn}
Z.~Yang, L.~Li, K.~Lin, J.~Wang, C.-C. Lin, Z.~Liu, and L.~Wang, ``The dawn of lmms: Preliminary explorations with gpt-4v (ision),'' \emph{arXiv preprint arXiv:2309.17421}, vol.~9, no.~1, p.~1, 2023.

\bibitem{comanici2025gemini}
G.~Comanici, E.~Bieber, M.~Schaekermann, I.~Pasupat, N.~Sachdeva, I.~Dhillon, M.~Blistein, O.~Ram, D.~Zhang, E.~Rosen \emph{et~al.}, ``Gemini 2.5: Pushing the frontier with advanced reasoning, multimodality, long context, and next generation agentic capabilities,'' \emph{arXiv preprint arXiv:2507.06261}, 2025.

\bibitem{reid2024gemini}
M.~Reid, N.~Savinov, D.~Teplyashin, D.~Lepikhin, T.~Lillicrap, J.-b. Alayrac, R.~Soricut, A.~Lazaridou, O.~Firat, J.~Schrittwieser \emph{et~al.}, ``Gemini 1.5: Unlocking multimodal understanding across millions of tokens of context,'' \emph{arXiv preprint arXiv:2403.05530}, 2024.

\bibitem{10721284}
J.~Liu, S.~Chen, X.~He, L.~Guo, X.~Zhu, W.~Wang, and J.~Tang, ``Valor: Vision-audio-language omni-perception pretraining model and dataset,'' \emph{IEEE Transactions on Pattern Analysis and Machine Intelligence}, vol.~47, no.~2, pp. 708--724, 2025.

\bibitem{10670217}
W.~Wu, X.~Wang, H.~Luo, J.~Wang, Y.~Yang, and W.~Ouyang, ``Cap4video++: Enhancing video understanding with auxiliary captions,'' \emph{IEEE Transactions on Pattern Analysis and Machine Intelligence}, vol.~47, no.~7, pp. 5223--5237, 2025.

\bibitem{10815073}
P.~Jin, H.~Li, L.~Yuan, S.~Yan, and J.~Chen, ``Hierarchical banzhaf interaction for general video-language representation learning,'' \emph{IEEE Transactions on Pattern Analysis and Machine Intelligence}, vol.~47, no.~3, pp. 2125--2139, 2025.

\bibitem{10839067}
X.~Wang, J.~Wu, Z.~Lin, F.~Zhang, D.~Zhang, and L.~Nie, ``Video dataflywheel: Resolving the impossible data trinity in video-language understanding,'' \emph{IEEE Transactions on Pattern Analysis and Machine Intelligence}, vol.~47, no.~4, pp. 2912--2923, 2025.

\bibitem{lin2023video}
B.~Lin, B.~Zhu, Y.~Ye, M.~Ning, P.~Jin, and L.~Yuan, ``Video-llava: Learning united visual representation by alignment before projection,'' \emph{arXiv preprint arXiv:2311.10122}, 2023.

\bibitem{zhou2024streaming}
X.~Zhou, A.~Arnab, S.~Buch, S.~Yan, A.~Myers, X.~Xiong, A.~Nagrani, and C.~Schmid, ``Streaming dense video captioning,'' in \emph{Proceedings of the IEEE/CVF Conference on Computer Vision and Pattern Recognition}, 2024, pp. 18\,243--18\,252.

\bibitem{xiong2025streaming}
H.~Xiong, Z.~Yang, J.~Yu, Y.~Zhuge, L.~Zhang, J.~Zhu, and H.~Lu, ``Streaming video understanding and multi-round interaction with memory-enhanced knowledge,'' \emph{arXiv preprint arXiv:2501.13468}, 2025.

\bibitem{zellers2022merlot}
R.~Zellers, J.~Lu, X.~Lu, Y.~Yu, Y.~Zhao, M.~Salehi, A.~Kusupati, J.~Hessel, A.~Farhadi, and Y.~Choi, ``Merlot reserve: Neural script knowledge through vision and language and sound,'' in \emph{Proceedings of the IEEE/CVF Conference on Computer Vision and Pattern Recognition}, 2022, pp. 16\,375--16\,387.

\bibitem{gao2023livechat}
J.~Gao, Y.~Lian, Z.~Zhou, Y.~Fu, and B.~Wang, ``Livechat: A large-scale personalized dialogue dataset automatically constructed from live streaming,'' \emph{arXiv preprint arXiv:2306.08401}, 2023.

\bibitem{grauman2022ego4d}
K.~Grauman, A.~Westbury, E.~Byrne, Z.~Chavis, A.~Furnari, R.~Girdhar, J.~Hamburger, H.~Jiang, M.~Liu, X.~Liu \emph{et~al.}, ``Ego4d: Around the world in 3,000 hours of egocentric video,'' in \emph{Proceedings of the IEEE/CVF Conference on Computer Vision and Pattern Recognition}, 2022, pp. 18\,995--19\,012.

\bibitem{fu2025vispeak}
S.~Fu, Q.~Yang, Y.-M. Li, Y.-X. Peng, K.-Y. Lin, X.~Wei, J.-F. Hu, X.~Xie, and W.-S. Zheng, ``Vispeak: Visual instruction feedback in streaming videos,'' \emph{arXiv preprint arXiv:2503.12769}, 2025.

\bibitem{kang2025open}
H.~Kang, Y.~Park, Y.~Yoo, Y.~Choi, and S.~J. Kim, ``Open-ended hierarchical streaming video understanding with vision language models,'' in \emph{Proceedings of the IEEE/CVF International Conference on Computer Vision}, 2025, pp. 20\,715--20\,725.

\bibitem{yang2025streamagent}
H.~Yang, F.~Tang, L.~Zhao, X.~An, M.~Hu, H.~Li, X.~Zhuang, Y.~Lu, X.~Zhang, A.~Swikir \emph{et~al.}, ``Streamagent: Towards anticipatory agents for streaming video understanding,'' \emph{arXiv preprint arXiv:2508.01875}, 2025.

\bibitem{qianlearning}
J.~Qian, H.~Du, G.~Nan, S.~Huang, J.~Yu, H.~Wang, J.~Chen, M.~Cai, M.~Yang, J.~Li \emph{et~al.}, ``Learning to respond: A large-scale benchmark and progressive learning framework for trigger-centric online video understanding.''

\bibitem{kim2025egospeak}
J.~Kim, M.-S. Kim, J.~Chung, J.~Cho, J.~Kim, S.~Kim, G.~Sim, and Y.~Yu, ``Egospeak: learning when to speak for egocentric conversational agents in the wild,'' in \emph{Findings of the Association for Computational Linguistics: NAACL 2025}, 2025, pp. 2990--3005.

\bibitem{mun2019streamlined}
J.~Mun, L.~Yang, Z.~Ren, N.~Xu, and B.~Han, ``Streamlined dense video captioning,'' in \emph{Proceedings of the IEEE/CVF conference on computer vision and pattern recognition}, 2019, pp. 6588--6597.

\bibitem{yan2026proact}
W.~Yan, Y.~Dai, Q.~Ran, H.~Li, W.~Lin, H.~Liao, X.~Xie, T.~Jin, and J.~Lian, ``Proact-vl: A proactive videollm for real-time ai companions,'' \emph{arXiv preprint arXiv:2603.03447}, 2026.

\bibitem{azad2026streamready}
S.~Azad, V.~Vineet, and Y.~S. Rawat, ``Streamready: Learning what to answer and when in long streaming videos,'' \emph{arXiv preprint arXiv:2603.08620}, 2026.

\bibitem{tian2026roma}
X.~Tian, W.~Li, B.~Xu, H.~Dong, Y.~Wang, and H.~Shen, ``Roma: Real-time omni-multimodal assistant with interactive streaming understanding,'' \emph{arXiv preprint arXiv:2601.10323}, 2026.

\bibitem{kim2026stride}
J.~Kim, H.~Lee, J.~M. Rehg, M.~Kim, and Y.~M. Ro, ``Stride: When to speak meets sequence denoising for streaming video understanding,'' \emph{arXiv preprint arXiv:2603.27593}, 2026.

\bibitem{zheng2026garde}
Y.~Zheng, X.~Ding, Y.~Yang, S.~Jiang, H.~Wu, Q.~Zhang, W.~Wang, T.~Cao, and Y.~Liu, ``Em-garde: A propose-match framework for proactive streaming video understanding,'' \emph{arXiv preprint arXiv:2603.19054}, 2026.

\bibitem{yao2025timechat}
L.~Yao, Y.~Li, Y.~Wei, L.~Li, S.~Ren, Y.~Liu, K.~Ouyang, L.~Wang, S.~Li, S.~Li \emph{et~al.}, ``Timechat-online: 80\% visual tokens are naturally redundant in streaming videos,'' in \emph{Proceedings of the 33rd ACM International Conference on Multimedia}, 2025, pp. 10\,807--10\,816.

\bibitem{chen2025livecc}
J.~Chen, Z.~Zeng, Y.~Lin, W.~Li, Z.~Ma, and M.~Z. Shou, ``Livecc: Learning video llm with streaming speech transcription at scale,'' in \emph{Proceedings of the Computer Vision and Pattern Recognition Conference}, 2025, pp. 29\,083--29\,095.

\bibitem{zhang2025eyes}
Y.~Zhang, C.~Shi, Y.~Wang, and S.~Yang, ``Eyes wide open: Ego proactive video-llm for streaming video,'' \emph{arXiv preprint arXiv:2510.14560}, 2025.

\bibitem{zhang2025proactive}
Y.~Zhang, X.~L. Dong, Z.~Lin, A.~Madotto, A.~Kumar, B.~Damavandi, J.~Chai, and S.~Moon, ``Proactive assistant dialogue generation from streaming egocentric videos,'' \emph{arXiv preprint arXiv:2506.05904}, 2025.

\bibitem{yang2025assistpda}
Z.~Yang, C.~Gao, J.~Liu, P.~Wu, G.~Pang, and M.~Z. Shou, ``Assistpda: An online video surveillance assistant for video anomaly prediction, detection, and analysis,'' \emph{arXiv preprint arXiv:2503.21904}, 2025.

\bibitem{xia2025streaming}
J.~Xia, P.~Chen, M.~Zhang, X.~Sun, and K.~Zhou, ``Streaming video instruction tuning,'' \emph{arXiv preprint arXiv:2512.21334}, 2025.

\bibitem{wang2025mmduet2}
Y.~Wang, S.~Liu, D.~Wang, N.~Xu, G.~Wan, H.~Zhang, and D.~Zhao, ``Mmduet2: Enhancing proactive interaction of video mllms with multi-turn reinforcement learning,'' \emph{arXiv preprint arXiv:2512.06810}, 2025.

\bibitem{liu2026thinking}
Z.~Liu, L.~Guo, H.~Li, R.~Zhen, X.~He, R.~Ji, X.~Ren, Y.~Zhang, H.~Lu, and J.~Liu, ``Thinking in streaming video,'' \emph{arXiv preprint arXiv:2603.12938}, 2026.

\bibitem{chen2026streamingclaw}
J.~Chen, Z.~Chen, C.~Du, M.~He, W.~He, H.~Li, Q.~Li, Z.~Liu, H.~Ma, X.~Pan \emph{et~al.}, ``Streamingclaw technical report,'' \emph{arXiv preprint arXiv:2603.22120}, 2026.

\bibitem{zhang2026querystream}
K.~Zhang, Z.~Yang, B.~Wang, S.~Qian, and C.~Xu, ``Querystream: Advancing streaming video understanding with query-aware pruning and proactive response,'' in \emph{The Fourteenth International Conference on Learning Representations}, 2026.

\bibitem{vaswani2017attention}
A.~Vaswani, N.~Shazeer, N.~Parmar, J.~Uszkoreit, L.~Jones, A.~N. Gomez, {\L}.~Kaiser, and I.~Polosukhin, ``Attention is all you need,'' \emph{Advances in neural information processing systems}, vol.~30, 2017.

\bibitem{team2024gemini}
G.~Team, P.~Georgiev, V.~I. Lei, R.~Burnell, L.~Bai, A.~Gulati, G.~Tanzer, D.~Vincent, Z.~Pan, S.~Wang \emph{et~al.}, ``Gemini 1.5: Unlocking multimodal understanding across millions of tokens of context,'' \emph{arXiv preprint arXiv:2403.05530}, 2024.

\bibitem{wang2025internvideo2}
Y.~Wang, X.~Li, Z.~Yan, Y.~He, J.~Yu, X.~Zeng, C.~Wang, C.~Ma, H.~Huang, J.~Gao \emph{et~al.}, ``Internvideo2. 5: Empowering video mllms with long and rich context modeling,'' \emph{arXiv preprint arXiv:2501.12386}, 2025.

\bibitem{yao2024minicpm}
Y.~Yao, T.~Yu, A.~Zhang, C.~Wang, J.~Cui, H.~Zhu, T.~Cai, H.~Li, W.~Zhao, Z.~He \emph{et~al.}, ``Minicpm-v: A gpt-4v level mllm on your phone,'' \emph{arXiv preprint arXiv:2408.01800}, 2024.

\bibitem{bai2025qwen2}
S.~Bai, K.~Chen, X.~Liu, J.~Wang, W.~Ge, S.~Song, K.~Dang, P.~Wang, S.~Wang, J.~Tang \emph{et~al.}, ``Qwen2. 5-vl technical report,'' \emph{arXiv preprint arXiv:2502.13923}, 2025.

\bibitem{qian2025dispider}
R.~Qian, S.~Ding, X.~Dong, P.~Zhang, Y.~Zang, Y.~Cao, D.~Lin, and J.~Wang, ``Dispider: Enabling video llms with active real-time interaction via disentangled perception, decision, and reaction,'' in \emph{Proceedings of the Computer Vision and Pattern Recognition Conference}, 2025, pp. 24\,045--24\,055.

\bibitem{lin2024streamingbench}
J.~Lin, Z.~Fang, C.~Chen, Z.~Wan, F.~Luo, P.~Li, Y.~Liu, and M.~Sun, ``Streamingbench: Assessing the gap for mllms to achieve streaming video understanding,'' \emph{arXiv preprint arXiv:2411.03628}, 2024.

\bibitem{li2025ovo}
Y.~Li, J.~Niu, Z.~Miao, C.~Ge, Y.~Zhou, Q.~He, X.~Dong, H.~Duan, S.~Ding, R.~Qian \emph{et~al.}, ``Ovo-bench: How far is your video-llms from real-world online video understanding?'' \emph{arXiv preprint arXiv:2501.05510}, 2025.

\bibitem{huang2025online}
Z.~Huang, X.~Li, J.~Li, J.~Wang, X.~Zeng, C.~Liang, T.~Wu, X.~Chen, L.~Li, and L.~Wang, ``Online video understanding: Ovbench and videochat-online,'' in \emph{Proceedings of the Computer Vision and Pattern Recognition Conference}, 2025, pp. 3328--3338.

\bibitem{wu2024longvideobench}
H.~Wu, D.~Li, B.~Chen, and J.~Li, ``Longvideobench: A benchmark for long-context interleaved video-language understanding,'' \emph{Advances in Neural Information Processing Systems}, vol.~37, pp. 28\,828--28\,857, 2024.

\bibitem{fu2025video}
C.~Fu, Y.~Dai, Y.~Luo, L.~Li, S.~Ren, R.~Zhang, Z.~Wang, C.~Zhou, Y.~Shen, M.~Zhang \emph{et~al.}, ``Video-mme: The first-ever comprehensive evaluation benchmark of multi-modal llms in video analysis,'' in \emph{Proceedings of the Computer Vision and Pattern Recognition Conference}, 2025, pp. 24\,108--24\,118.

\bibitem{chen2024expanding}
Z.~Chen, W.~Wang, Y.~Cao, Y.~Liu, Z.~Gao, E.~Cui, J.~Zhu, S.~Ye, H.~Tian, Z.~Liu \emph{et~al.}, ``Expanding performance boundaries of open-source multimodal models with model, data, and test-time scaling,'' \emph{arXiv preprint arXiv:2412.05271}, 2024.

\bibitem{cai2024internlm2}
Z.~Cai, M.~Cao, H.~Chen, K.~Chen, K.~Chen, X.~Chen, X.~Chen, Z.~Chen, Z.~Chen, P.~Chu \emph{et~al.}, ``Internlm2 technical report,'' \emph{arXiv preprint arXiv:2403.17297}, 2024.

\bibitem{zheng2023judging}
L.~Zheng, W.-L. Chiang, Y.~Sheng, S.~Zhuang, Z.~Wu, Y.~Zhuang, Z.~Lin, Z.~Li, D.~Li, E.~Xing, H.~Zhang, J.~E. Gonzalez, and I.~Stoica, ``Judging {LLM}-as-a-judge with {MT}-bench and chatbot arena,'' in \emph{Advances in Neural Information Processing Systems (NeurIPS)}, 2023.

\bibitem{liu2023geval}
Y.~Liu, D.~Iter, Y.~Xu, S.~Wang, R.~Xu, and C.~Zhu, ``{G-Eval}: {NLG} evaluation using {GPT-4} with better human alignment,'' in \emph{Proceedings of the 2023 Conference on Empirical Methods in Natural Language Processing (EMNLP)}, 2023.

\bibitem{chiang2023alternative}
C.-H. Chiang and H.-y. Lee, ``Can large language models be an alternative to human evaluations?'' in \emph{Proceedings of the 61st Annual Meeting of the Association for Computational Linguistics (ACL)}, 2023.

\bibitem{11168273}
L.~Hong, Z.~Liu, W.~Chen, C.~Tan, Y.~Feng, X.~Zhou, P.~Guo, J.~Li, Z.~Chen, S.~Gao, W.~Zhang, and W.~Zhang, ``Lvos: A benchmark for large-scale long-term video object segmentation,'' \emph{IEEE Transactions on Pattern Analysis and Machine Intelligence}, vol.~48, no.~1, pp. 946--961, 2026.

\bibitem{11097075}
S.~Yang, W.~Yu, W.~Yang, X.~Liu, H.~Tan, L.~Lan, and N.~Xiao, ``Wildvideo: Benchmarking lmms for understanding video-language interaction,'' \emph{IEEE Transactions on Pattern Analysis and Machine Intelligence}, vol.~47, no.~10, pp. 9330--9344, 2025.

\bibitem{chen2024far}
Z.~Chen, W.~Wang, H.~Tian, S.~Ye, Z.~Gao, E.~Cui, W.~Tong, K.~Hu, J.~Luo, Z.~Ma \emph{et~al.}, ``How far are we to gpt-4v? closing the gap to commercial multimodal models with open-source suites,'' \emph{arXiv preprint arXiv:2404.16821}, 2024.

\bibitem{li2024llama}
Y.~Li, C.~Wang, and J.~Jia, ``Llama-vid: An image is worth 2 tokens in large language models,'' in \emph{European Conference on Computer Vision}.\hskip 1em plus 0.5em minus 0.4em\relax Springer, 2024, pp. 323--340.

\bibitem{li2024llava}
B.~Li, Y.~Zhang, D.~Guo, R.~Zhang, F.~Li, H.~Zhang, K.~Zhang, P.~Zhang, Y.~Li, Z.~Liu \emph{et~al.}, ``Llava-onevision: Easy visual task transfer,'' \emph{arXiv preprint arXiv:2408.03326}, 2024.

\bibitem{Qwen2VL}
P.~Wang, S.~Bai, S.~Tan, S.~Wang, Z.~Fan, J.~Bai, K.~Chen, X.~Liu, J.~Wang, W.~Ge, Y.~Fan, K.~Dang, M.~Du, X.~Ren, R.~Men, D.~Liu, C.~Zhou, J.~Zhou, and J.~Lin, ``Qwen2-vl: Enhancing vision-language model's perception of the world at any resolution,'' \emph{arXiv preprint arXiv:2409.12191}, 2024.

\bibitem{huang2024lita}
D.-A. Huang, S.~Liao, S.~Radhakrishnan, H.~Yin, P.~Molchanov, Z.~Yu, and J.~Kautz, ``Lita: Language instructed temporal-localization assistant,'' in \emph{European Conference on Computer Vision}.\hskip 1em plus 0.5em minus 0.4em\relax Springer, 2024, pp. 202--218.

\bibitem{huang2024vtimellm}
B.~Huang, X.~Wang, H.~Chen, Z.~Song, and W.~Zhu, ``Vtimellm: Empower llm to grasp video moments,'' in \emph{Proceedings of the IEEE/CVF Conference on Computer Vision and Pattern Recognition}, 2024, pp. 14\,271--14\,280.

\bibitem{song2024moviechat}
E.~Song, W.~Chai, G.~Wang, Y.~Zhang, H.~Zhou, F.~Wu, H.~Chi, X.~Guo, T.~Ye, Y.~Zhang \emph{et~al.}, ``Moviechat: From dense token to sparse memory for long video understanding,'' in \emph{Proceedings of the IEEE/CVF Conference on Computer Vision and Pattern Recognition}, 2024, pp. 18\,221--18\,232.

\bibitem{rumelhart1985learning}
D.~E. Rumelhart, G.~E. Hinton, and R.~J. Williams, ``Learning internal representations by error propagation,'' Tech. Rep., 1985.

\bibitem{hochreiter1997long}
S.~Hochreiter and J.~Schmidhuber, ``Long short-term memory,'' \emph{Neural computation}, vol.~9, no.~8, pp. 1735--1780, 1997.

\end{thebibliography}

\vfill

\end{document}